\documentclass{article}

\usepackage[main, final]{neurips_2026}

\usepackage[utf8]{inputenc}
\usepackage[T1]{fontenc}
\usepackage{hyperref}
\usepackage{url}
\usepackage{booktabs}
\usepackage{multirow}
\usepackage{amsfonts}
\usepackage{amssymb}
\usepackage{amsmath}
\usepackage{nicefrac}
\usepackage{microtype}
\usepackage[table]{xcolor}
\usepackage{graphicx}
\usepackage{float}
\usepackage{wrapfig}
\usepackage{subcaption}
\usepackage{mathtools}
\usepackage{amsthm}
\usepackage[ruled,linesnumbered,noend]{algorithm2e}
\usepackage[capitalize,noabbrev]{cleveref}
\usepackage{tikz}
\usetikzlibrary{arrows.meta, positioning, shapes.geometric, calc, fit, backgrounds}

\SetAlgoLined
\SetKwInput{KwInput}{Input}
\SetKwInput{KwOutput}{Output}
\SetKwComment{Comment}{$\triangleright$~}{}

\SetCommentSty{commentfont}
\SetKwFor{ForAll}{for all}{do}{}
\DontPrintSemicolon
\SetAlCapSty{textbf}
\SetAlCapNameSty{textbf}

\SetAlgoCaptionSeparator{.~}

\usepackage{enumitem}

\usepackage{fancyvrb}
\usepackage{fvextra}
\DefineVerbatimEnvironment{Verbatim}{Verbatim}{breaklines=true, breakanywhere=true}

\definecolor{ysdarkpurple}{HTML}{4E2399}
\definecolor{ysshallowpurple}{HTML}{E6DBFF}
\definecolor{ysdarkred}{HTML}{8c2824}
\definecolor{ysshallowred}{HTML}{F8D7D7}
\definecolor{ysdarkblue}{HTML}{005E99}
\definecolor{ysshallowblue}{HTML}{CCEBFF}
\definecolor{ysdarkgrey}{HTML}{333333}
\definecolor{ysshallowgrey}{HTML}{E5E5E5}
\definecolor{ysultralight}{HTML}{F5F5F5}
\definecolor{ysdarkgreen}{HTML}{2D7A3E}
\definecolor{ysshallowgreen}{HTML}{D4F0DD}

\graphicspath{{figures/}}

\title{TACT: Mitigating Agent Drift via Activation Steering}
\title{TACT: Mitigating Overthinking and Overacting via Activation Steering}
\title{TACT: Mitigating Overthinking and Overacting in Coding Agents via Activation Steering}

\author{%
  \textbf{Yuan Sui}$^{1}$, \textbf{Yulin Chen}$^{1}$, \textbf{Yibo Li}$^{1}$, \textbf{Xue Jiang}$^{2}$, \textbf{Yufei He}$^{1}$, \textbf{Yihong Dong}$^{2}$ \\
  \textbf{Xiaoxin He}$^{3}$, \textbf{Tianyu Gao}$^{1}$, \textbf{Bryan Hooi}$^{1}$ \\
  $^{1}$National University of Singapore \quad $^{2}$Peking University \quad $^{3}$Meta \\
}

\begin{document}

\maketitle

\begin{abstract}

When language model agents tackle complex software engineering tasks, they often degrade over long trajectories, which we define as \textit{agent drift}. We focus on two recurring failure modes \textit{overthinking} and \textit{overacting}, i.e., where the agent repeatedly reasons over information it already has, and where it issues tool calls without integrating recent observations or acquiring new evidence. 
In this paper, we introduce \textbf{TACT} (\textbf{T}hink-\textbf{A}ct \textbf{C}alibration via activation s\textbf{T}eering), to detect and mitigate agent drift in the residual stream before it surfaces as a behavioral failure.
In specific, we label trajectory steps as overthinking, overacting, or calibrated, and find that their hidden states can separate linearly along two \textit{drift axes}, pointing from calibrated behavior toward each failure mode (AUC $\approx$ 0.9). To mitigate agent drift, we project each step's activation onto these axes at test time and pull drifted ones back toward the calibrated region.
Experiments show that \textsc{TACT} outperforms unsteered baselines across SWE-bench Verified, Terminal-Bench 2.0, and CLAW-Eval, lifting average resolve rate by $+5.8$ pp on Qwen3.5-27B and $+4.8$ pp on Gemma-4-26B-A4B-it while cutting steps-to-resolve by up to $26\%$.
These gains frame agent drift as a steerable direction in the residual stream, and position TACT as a viable handle for reliable long-horizon agents.


\end{abstract}

\section{Introduction}
\label{sec:introduction}


Software engineering benchmarks such as SWE-bench~\citep{jimenez2024swebench} have become a central testbed for evaluating autonomous coding agents.
Unlike single-turn code generation, these benchmarks require agents to operate over long horizons: they must navigate large repositories, localize bugs, and produce correct patches.
A successful trajectory therefore depends not only on whether the underlying language model can reason or generate code, but also on whether the agent can continuously allocate its computation between \textit{thinking} and \textit{acting}, which is empirically fragile~\citep{wang2025toa}.
As trajectories grow to dozens or hundreds of steps~\citep{yang2024sweagent}, agents often degrade over time, which we refer to as \textit{agent drift}. We focus on two failure modes \textit{overthinking} and \textit{overacting}, i.e., where the agent repeatedly reasons over information it already has, and where it issues tool calls without integrating recent observations or acquiring new evidence.

These failures matter because they affect both task success and efficiency.
Overthinking wastes context and delays useful actions, and also can cause failure: when the agent keeps reasoning from the same evidence, it may reinforce a wrong hypothesis~\citep{zhao2025test}, postpone the tool call that would disconfirm it, and run out of steps or context before finding the right fix~\citep{cuadron2025overthinking}. In contrast, overacting wastes tool calls by repeating searches, re-reading files, or running commands without first using the information already observed; tools should be called only when they are needed to obtain new evidence~\citep{wang2025toa}. It can also cause failure by flooding the trajectory with redundant observations, making it harder to track which evidence matters and which hypothesis should be updated.
Both patterns reduce agent performance, but they are hard to diagnose from final outcomes alone: two failed trajectories can look the same even if one failed by premature action and the other by excessive deliberation. This suggests coding agent evaluation should move beyond task-level success and check whether each step is well-calibrated to the agent's current information state~\citep{yehudai2025agenteval}.

A growing body of work studies such failures behaviorally~\citep{cuadron2025overthinking, wang2025toa, yue2025dontoverthink, rath2026agentdrift} and mitigates them through mechanisms outside the model.
Reflexion~\citep{shinn2023reflexion} and Self-Refine~\citep{madaan2023selfrefine} ask the model to critique and revise its own outputs.
Tree-of-Thoughts~\citep{yao2023tot} searches over alternative reasoning paths.
Meta-reasoner~\citep{sui2025metareasoner} uses an external controller to select reasoning strategies.
These approaches are useful, but they treat the model as a black box: drift is observed only after it manifests in reasoning context or tool use, and interventions are applied through prompts, search, or orchestration.
This leaves open a lower-level question: \textbf{Can agent drift be detected and mitigated at test time, before it fully surfaces as a behavioral failure?}


\begin{figure}[t]
  \centering
  \includegraphics[width=\linewidth]{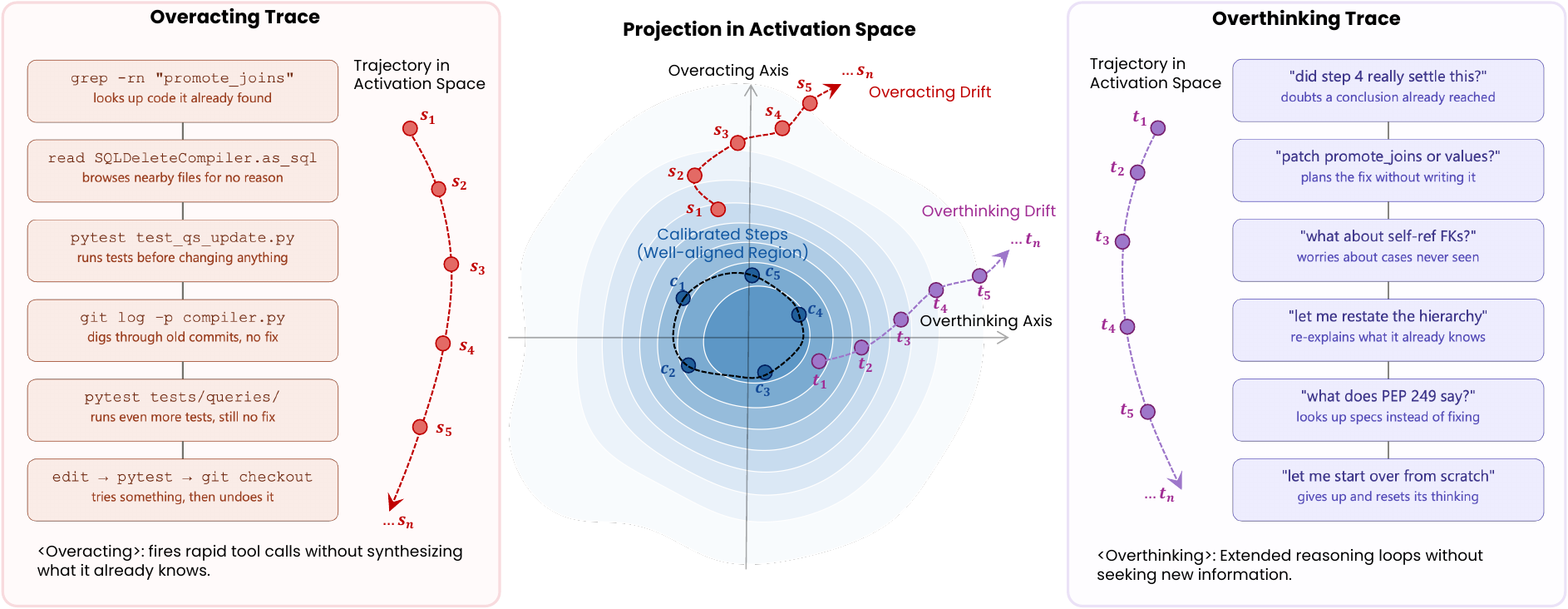}
  \caption{Two failure modes of a coding agent, mapped into activation space. The left and right panels show concrete steps from an overacting trajectory (left) and an overthinking trajectory (right). The center panel projects each step's hidden state at the \texttt{</think>} token onto our two drift axes: calibrated steps cluster in the central band, while overthinking and overacting trajectories drift outward in opposite directions. Both failure modes are linearly separable from calibrated activations along their respective axes, motivating the dual-axis steering used at test time (Section~\ref{sec:steering}).}
  \label{fig:demonstration}
  \vspace{-5mm}
\end{figure}

This question is motivated by representation engineering (RepE)~\citep{zou2023repeng}, showing that high-level model behaviors are often linearly represented in transformer residual streams.
Properties such as refusal~\citep{arditi2024refusal}, truthfulness~\citep{li2023iti}, and style or persona~\citep{zou2023repeng} can be captured by contrastive activation directions and steered at inference time. The closest precedent is the assistant axis of \citet{lu2026assistant}, which stabilizes an instruction-tuned model's assistant-like behavior over long contexts.
However, existing activation-steering results mostly target behaviors that can be identified from a single model response, such as refusal, truthfulness, persona, or response style.
Long-horizon agents pose a different problem: whether a step is overthinking or overacting depends not only on the step itself, but also on what the agent has already observed and what information is still needed. It is therefore unclear whether such context-dependent failures can be captured by a simple linear direction in activation space as shown in Figure~\ref{fig:demonstration}.

\begin{figure}[t]
  \centering
  \resizebox{\textwidth}{!}{%
  \begin{tikzpicture}[
    font=\small,
    >={Stealth[length=2mm,width=1.6mm]},
    stage/.style={draw=black!70, thick, rounded corners=2pt, minimum width=5.4cm, minimum height=3cm, fill=white},
    title/.style={font=\footnotesize\bfseries, text=black!85},
    sub/.style={font=\scriptsize, text=black!65, align=center},
    flow/.style={->, very thick, black!70},
    otdot/.style={circle, fill=ysdarkred, inner sep=1.4pt},
    oadot/.style={circle, fill=ysdarkpurple, inner sep=1.4pt},
    caldot/.style={circle, fill=ysdarkblue, inner sep=1.4pt},
    stepbox/.style={draw=black!45, rounded corners=1pt, fill=white,
                    minimum width=0.5cm, minimum height=0.4cm, inner sep=1pt, font=\tiny},
  ]

  \node[stage] (s1) {};
  \node[title, anchor=south] at ([yshift=2pt]s1.north) {Stage 1: Trajectory annotation};
  \node[sub, anchor=north] at ([yshift=-4pt]s1.north) {Agent steps $s_1, \dots, s_N$ \\ labeled by LLM-as-judge};

  \node[stepbox, fill=ysshallowred]    (b1) at ($(s1.center)+(-2.00, 0.10)$) {$s_1$};
  \node[stepbox, fill=ysshallowred]    (b2) at ($(s1.center)+(-1.50, 0.10)$) {$s_2$};
  \node[stepbox, fill=ysshallowblue]   (b3) at ($(s1.center)+(-1.00, 0.10)$) {$s_3$};
  \node[stepbox, fill=ysshallowpurple] (b4) at ($(s1.center)+(-0.50, 0.10)$) {$s_4$};
  \node[stepbox, fill=ysshallowpurple] (b5) at ($(s1.center)+( 0.00, 0.10)$) {$s_5$};
  \node[stepbox, fill=ysshallowblue]   (b6) at ($(s1.center)+( 0.50, 0.10)$) {$s_6$};
  \node[stepbox, fill=ysshallowred]    (b7) at ($(s1.center)+( 1.00, 0.10)$) {$s_7$};
  \node[stepbox, fill=ysshallowblue]   (b8) at ($(s1.center)+( 1.50, 0.10)$) {$s_8$};
  \node[stepbox, fill=ysshallowred]    (b9) at ($(s1.center)+( 2.00, 0.10)$) {$s_9$};

  \node[font=\tiny\bfseries, text=ysdarkred]    at ([yshift=-9pt]b1.south) {OT};
  \node[font=\tiny\bfseries, text=ysdarkred]    at ([yshift=-9pt]b2.south) {OT};
  \node[font=\tiny\bfseries, text=ysdarkblue]   at ([yshift=-9pt]b3.south) {CAL};
  \node[font=\tiny\bfseries, text=ysdarkpurple] at ([yshift=-9pt]b4.south) {OA};
  \node[font=\tiny\bfseries, text=ysdarkpurple] at ([yshift=-9pt]b5.south) {OA};
  \node[font=\tiny\bfseries, text=ysdarkblue]   at ([yshift=-9pt]b6.south) {CAL};
  \node[font=\tiny\bfseries, text=ysdarkred]    at ([yshift=-9pt]b7.south) {OT};
  \node[font=\tiny\bfseries, text=ysdarkblue]   at ([yshift=-9pt]b8.south) {CAL};
  \node[font=\tiny\bfseries, text=ysdarkred]    at ([yshift=-9pt]b9.south) {OT};

  \node[font=\tiny, anchor=south] at ([yshift=4pt]s1.south)
    {{\color{ysdarkred}$\blacksquare$}\,\textbf{OT}~overthinking \;
     {\color{ysdarkpurple}$\blacksquare$}\,\textbf{OA}~overacting \;
     {\color{ysdarkblue}$\blacksquare$}\,\textbf{CAL}~calibrated};

  \node[stage, right=1.1cm of s1] (s2) {};
  \node[title, anchor=south] at ([yshift=2pt]s2.north) {Stage 2: Drift axes};
  \node[sub, anchor=north] at ([yshift=-4pt]s2.north) {Hidden states at \texttt{</think>} \\ projected to activation space};

  \begin{scope}[shift={(s2.center)}]
    \foreach \p in {(-0.05,0.05), (0.10,-0.10), (-0.15,-0.05), (0.05,0.18), (0.22,0.06), (-0.12,0.20), (0.0,-0.22), (0.15,-0.18)} {
      \node[caldot] at \p {};
    }
    \foreach \p in {(0.95,0.55), (1.05,0.40), (1.15,0.62), (0.85,0.45), (1.25,0.50)} {
      \node[otdot] at \p {};
    }
    \foreach \p in {(-0.95,-0.45), (-1.05,-0.58), (-0.85,-0.35), (-1.15,-0.50), (-0.90,-0.62)} {
      \node[oadot] at \p {};
    }
    \draw[->, thick, ysdarkred!85,    dashed] (0,0) -- (1.3, 0.65);
    \draw[->, thick, ysdarkpurple!85, dashed] (0,0) -- (-1.3,-0.55);
    \node[font=\tiny, text=ysdarkred]    at (1.55, 0.70) {$\hat{\mathbf{v}}_{\text{OT}}$};
    \node[font=\tiny, text=ysdarkpurple] at (-1.55,-0.62) {$\hat{\mathbf{v}}_{\text{OA}^\perp}$};
  \end{scope}

  \node[sub, anchor=south] at ([yshift=2pt]s2.south) {linear separation of OT, OA, CAL};

  \node[stage, right=1.1cm of s2] (s3) {};
  \node[title, anchor=south] at ([yshift=2pt]s3.north) {Stage 3: Dual-axis steering};
  \node[sub, anchor=north] at ([yshift=-4pt]s3.north) {Test-time intervention \\ at out-of-band activations};

  \begin{scope}[shift={(s3.center)}]
    \fill[ysdarkblue!12, rounded corners=2pt] (-1.5,-0.22) rectangle (1.5,0.22);
    \draw[ysdarkblue!55, thin, dashed] (-1.5, 0.22) -- (1.5, 0.22);
    \draw[ysdarkblue!55, thin, dashed] (-1.5,-0.22) -- (1.5,-0.22);
    \node[font=\tiny, text=ysdarkblue!90] at (0, 0) {calibrated band};

    \node[otdot] (p1) at (0.95, 0.65) {};
    \node[caldot] (p1b) at (0.85, 0.12) {};
    \draw[->, thick, ysdarkred!75] (p1) -- (p1b);

    \node[oadot] (p2) at (-1.05, -0.65) {};
    \node[caldot] (p2b) at (-0.85, -0.10) {};
    \draw[->, thick, ysdarkpurple!75] (p2) -- (p2b);
  \end{scope}

  \node[sub, anchor=south] at ([yshift=2pt]s3.south)
    {$\Rightarrow$ higher resolve rate, no retraining};

  \draw[flow] (s1.east) -- (s2.west);
  \draw[flow] (s2.east) -- (s3.west);

  \end{tikzpicture}%
  }
  \caption{\textbf{Pipeline overview.} Left to right: (1) we run an agent on a coding task and use an LLM-as-judge to label every step as \textcolor{ysdarkred}{\textbf{overthinking}} (OT), \textcolor{ysdarkpurple}{\textbf{overacting}} (OA), or \textcolor{ysdarkblue}{\textbf{calibrated}} (CAL); (2) we extract hidden states at the reasoning-action boundary (the \texttt{</think>} token) and compute contrastive drift axes that linearly separate failure modes from calibrated behavior; (3) at test time we apply dual-axis steering to nudge out-of-band activations back into the calibrated band without retraining.}
  \label{fig:pipeline}
  \vspace{-5mm}
\end{figure}
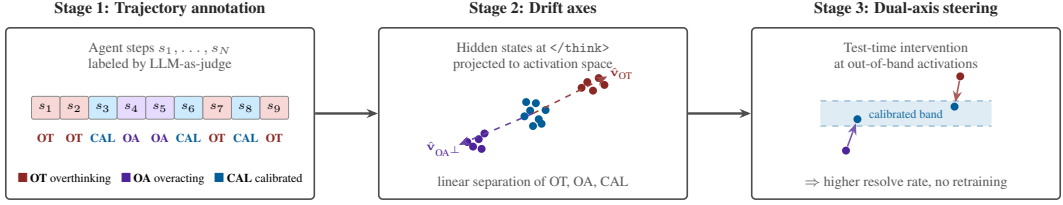

To this end, we propose \textbf{TACT} (\textbf{T}hink-\textbf{A}ct \textbf{C}alibration via activation s\textbf{T}eering), a three-stage pipeline to detect and mitigate agent drift in the residual stream before it surfaces as a behavioral failure:
\begin{enumerate}[leftmargin=1.2em,itemsep=1pt,topsep=2pt,parsep=0pt]
  \item \textbf{Annotate.} We design an LLM-as-judge pipeline~\citep{zhuge2024agentjudge} that labels each step of an agent trajectory as overthinking, overacting, or calibrated.
  The pipeline carries a validated rolling state, a compressed summary of the agent's accumulated knowledge audited by a dedicated validator with a reflection loop, so each step is judged in full context rather than in isolation.
    \item \textbf{Extract.} We collect hidden states at the \texttt{</think>} token for each labeled step and compute contrastive axes via the balanced mean-difference method~\citep{zou2023repeng}. For intervention, we use two failure-vs-calibrated axes, overthinking-vs-calibrated and overacting-vs-calibrated, and orthogonalize them via Gram-Schmidt~\citep{trefethen1997numerical} to enable independent steering. We additionally compute an overthinking-vs-overacting axis to analyze the three-class geometry.
  \item \textbf{Test-time Steering.} At test time, we project each step's hidden state onto the two axes; when a step drifts toward overthinking or overacting, we subtract a small step along the corresponding axis to pull the residual stream back toward the calibrated region, without fine-tuning or prompt changes~\citep{turner2023steering, panickssery2024contrastive}.
  Figure~\ref{fig:pipeline} illustrates the full pipeline.
\end{enumerate}

Our contributions:
(1) We formalize \textit{overthinking} and \textit{overacting} as agent's step-level miscalibration, and build a three-stage LLM-as-judge pipeline that labels each step as overthinking, overacting, or calibrated.
(2) We find that across transformer layers, the hidden states of annotated steps can form well-separable clusters (AUC $\approx$ 0.9); to analyze agent drift, we derive two \textit{drift axes} as the vectors from the centroid of the calibrated steps to the centroids of the overthinking and overacting steps.
(3) We demonstrate that test-time steering along these axes improves agent performance across SWE-bench Verified, Terminal-Bench 2.0, and CLAW-Eval, raising resolve rate by $+5.8$ pp on Qwen3.5-27B and $+4.8$ pp on Gemma-4-26B-A4B-it while reducing steps-to-resolve by up to $26\%$.

%
\section{Method}
\label{sec:method}

\subsection{Problem Formulation}
\label{sec:formulation}

Consider a coding agent fixing a bug in a large repository.
At each step it can make progress through internal reasoning (analyzing a stack trace, forming a hypothesis) or external interaction (running \texttt{grep -rn "QuerySet"}, executing \texttt{pytest}).
Either mode can drift: the agent may re-derive facts it already has, or call tools to retrieve information it already knows~\citep{wang2025toa}.

Let a trajectory $T=(s_1,\dots,s_N)$ be a sequence of steps, where $s_t=(r_t,a_t,o_t)$ consists of a reasoning block, an action, and an observation.
Let $\mathcal{K}_t$ denote the agent's \textit{information state} before step $t$: the facts, hypotheses, and observations accumulated from steps $1,\dots,t-1$.
We classify each step by whether it expands $\mathcal{K}_t$.
\textbf{Overthinking.} $r_t$ re-derives a conclusion already in $\mathcal{K}_t$, or spends reasoning on a question a simple tool call would resolve.
\textbf{Overacting.} $a_t$ returns an observation already in $\mathcal{K}_t$, or fires without adequately processing recent observations.
\textbf{Calibrated.} The step expands $\mathcal{K}_t$, either by deriving a new hypothesis or by retrieving previously unavailable information.

\begin{figure}[t]
  \centering
  \includegraphics[width=\linewidth]{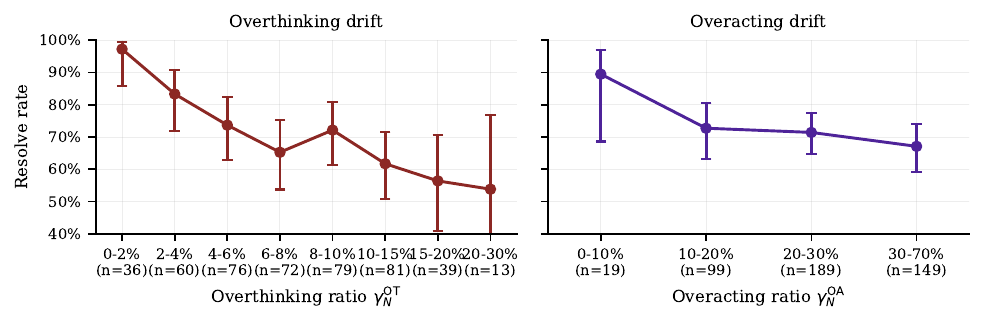}
  \caption{Resolve rate vs trajectory-level drift ratio on Qwen3.5-27B SWE-bench Verified, binned by $\gamma_N^{\text{OT}}$ (left) and $\gamma_N^{\text{OA}}$ (right). Error bars are Wilson 95\% confidence intervals; $n$ is the bin's trajectory count. Higher overthinking and overacting ratios both correlate with lower resolve rates.}
  \label{fig:drift_ratio_resolve}
  \vspace{-6mm}
\end{figure}

To quantify the trajectory-level agent drift, we aggregate the step-level labels into trajectory-level metrics, denoted as \textit{drift ratios} $\gamma_t$. Let $y_i \in \{\text{OT}, \text{OA}, \text{CAL}\}$ denote the label of step $i$. After $t$ steps, we define drift ratios $\gamma_t^{\text{OT}}$ and $\gamma_t^{\text{OA}}$ as the fractions of those $t$ steps labeled OT and OA:
\begin{equation}  
  \gamma_t^{\text{OT}} = \tfrac{1}{t}\sum_{i=1}^{t}\mathbf{1}[y_i = \text{OT}], \qquad
  \gamma_t^{\text{OA}} = \tfrac{1}{t}\sum_{i=1}^{t}\mathbf{1}[y_i = \text{OA}].
  \label{eq:gamma}
\end{equation}
We find that both $\gamma_N^{\text{OT}}$ and $\gamma_N^{\text{OA}}$ correlate with lower resolve rate on SWE-bench Verified (Figure~\ref{fig:drift_ratio_resolve}): an agent with higher $\gamma_N$ is less likely to fix the bug.
This tight link between drift and resolve rate motivates our approach: if step-level drift is detectable in the agent's activations, then correcting it during a rollout should raise the resolve rate, with no fine-tuning or prompt changes.

\subsection{Trajectory Annotation via LLM-as-Judge}
\label{sec:judge}

\begin{figure}[ht]
\centering
\includegraphics[width=\linewidth]{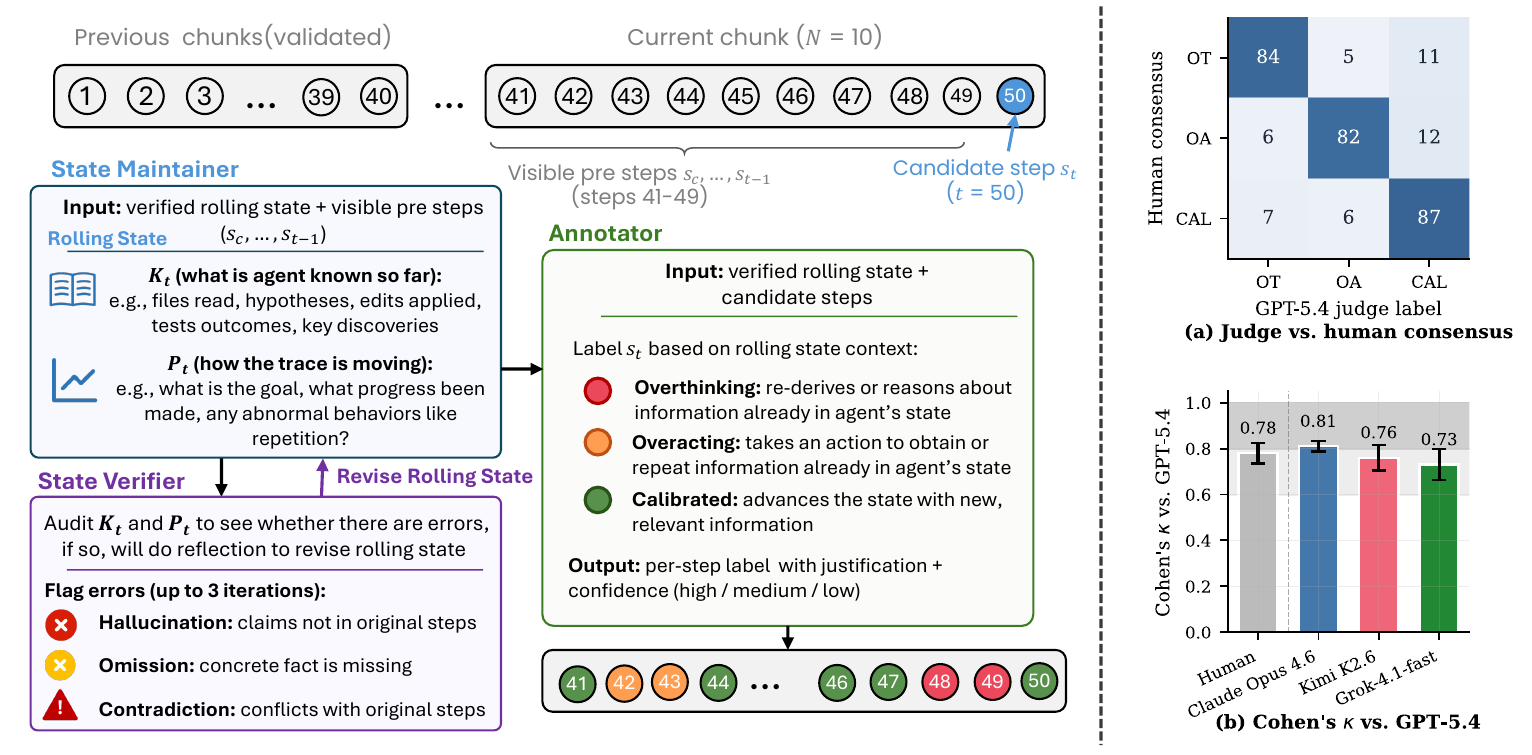}
\caption{Per-step annotation pipeline and judge reliability.
\textbf{Left}: the trajectory is split into chunks of $W$ steps.
For each chunk, a \textit{State Maintainer} updates the rolling state; a \textit{State Verifier} audits it against raw steps and triggers revisions when necessary; and an \textit{Annotator} labels each step using the verified rolling state from earlier chunks and the prefix within the current chunk.
\textbf{Right}: reliability analysis of the LLM-as-Judge.
(a) Confusion matrix between the judge and human consensus, showing recalls of $84\%/82\%/87\%$ for OT/OA/CAL.
(b) Cohen's $\kappa$ between the judge and four references (human consensus, Claude Opus~4.6, Kimi K2.6, and Grok-4.1-fast) falls in the substantial range ($0.73$--$0.81$).
Full reliability analysis appears in Appendix~\ref{app:judge_reliability}.}
\label{fig:judge}
\vspace{-5mm}
\end{figure}

To extract drift axes, we need labels for each trajectory step: was step $i$ overthinking, overacting, or calibrated? The challenge is to manage context efficiently. Judging step 50 requires knowing what the agent learned in steps 1--49; a stateless judge cannot tell ``re-reading a file already seen'' from ``reading a new file.'' The trajectory is divided into chunks of $W$ steps (we use $W{=}10$). Before annotating a chunk, we carry forward a verified state summarizing all previous chunks.
This state has two components:
\textit{KnownFacts} records concrete information the agent has observed so far, including files read, hypotheses under consideration, edits applied, test outcomes, and key discoveries.
\textit{Progress} records how the agent's behavior is evolving, including the current bug-fixing stage, how many steps have passed since KnownFacts last grew, and counts of re-read files, repeated actions, and cycled hypotheses.
For each chunk, we run three agents in sequence: a state maintainer, a state verifier, and an annotator. The overall pipeline is shown in Figure~\ref{fig:judge}.

\textbf{Stage 1: State Maintainer.}
Given the verified rolling state from previous chunks and the raw steps in the current chunk, the maintainer updates the rolling state. Its role is to summarize what the agent has newly learned or done in the current chunk. In particular, it updates KnownFacts with new observations, edits, and test results, and updates Progress with signals of stagnation or repetition.

\textbf{Stage 2: State Verifier.}
Because errors in the rolling state can accumulate over long trajectories, we use a separate verifier to audit the maintained state against the raw steps.
The verifier checks for three types of errors:
(1) \textit{hallucination}, where the state claims something not supported by the raw steps, such as listing a file as read when no step reads it;
(2) \textit{omission}, where a concrete fact from the raw steps is missing, such as leaving \texttt{edits\_made} empty after an edit step; and
(3) \textit{contradiction}, where the state conflicts with the step text, such as recording a test as passed when the observation shows failure.
If the verifier flags an error, the maintainer revises the state using the verifier's feedback, and the verifier audits the revised state again.

\textbf{Stage 3: Annotator.}
After verification, the annotator labels each step in the chunk as \textsc{overthinking}, \textsc{overacting}, or \textsc{calibrated}. Each label comes with a justification grounded in the rolling state and a confidence level (high, medium, low); we keep only high- and medium-confidence labels for axis extraction. Full system prompts and the rolling state schema are in Appendix~\ref{app:prompt}.

\subsection{Drift Axes Extraction}
\label{sec:axis}

Given per-step labels, we extract hidden states to build a geometric representation of agent behavior.
We focus on the \texttt{</think>} token as a canonical decision point, which marks the end of reasoning and the start of actions.
For each rollout, we run a single forward pass through the model and extract the hidden states of each \texttt{</think>} token under the causal mask. 
This yields a per-step activation tensor 
$\mathbf{H}_t \in \mathbb{R}^{L \times d}$,
where $L$ is the number of layers and $d$ is the hidden dimension.

Overthinking and overacting represent distinct failure modes, and we do not assume that they lie on a single shared direction relative to calibrated behavior.
We therefore construct separate contrastive directions for each failure mode against calibrated steps using the mean-difference method~\citep{zou2023repeng}.
In addition, we compute the OT-vs-OA direction as a diagnostic axis to characterize how the two failure modes are arranged in activation space.
For each problem, we compute the difference between the two class means and then average these differences across problems.
For an axis pointing from class $B$ to class $A$ at layer $\ell$, this is:
\begin{equation}
  \mathbf{v}_{A \text{-vs-} B}^{(\ell)} = \frac{1}{|J|} \sum_{j \in J} \left( \frac{1}{|S_{A}^{j}|} \sum_{t \in S_{A}^{j}} \mathbf{h}_t^{(\ell)} - \frac{1}{|S_{B}^{j}|} \sum_{t \in S_{B}^{j}} \mathbf{h}_t^{(\ell)} \right)
  \label{eq:axis}
\end{equation}
where $J$ is the set of problems containing both classes and $S_{A}^{j}$, $S_{B}^{j}$ are the steps of problem $j$ labeled $A$ and $B$.
Each axis is L2-normalized per layer. 
We define three axes accordingly: (1) $\mathbf{v}_{\text{OT-vs-CAL}}$, (2) $\mathbf{v}_{\text{OA-vs-CAL}}$ and (3) $\mathbf{v}_{\text{OT-vs-OA}}$.
We use the first two for TACT steering. The third axis $\mathbf{v}_{\text{OT-vs-OA}}$ points from one failure mode to the other rather than toward calibrated behavior, so steering along it would shift between two failure modes instead of moving the agent toward the CAL cluster. We therefore reserve $\mathbf{v}_{\text{OT-vs-OA}}$ for analyzing three-class geometry in Section~\ref{sec:axis_evidence}.
The two failure-mode axes are not designed to be orthogonal, so steering along one would interfere with steering along the other. To remove this interference, we fix one axis as the primary direction at each layer and orthogonalize the other against it via Gram-Schmidt~\citep{trefethen1997numerical}.
Our experiments find OT-vs-CAL the more separable of the two on held-out problems (Section~\ref{sec:axis_evidence}), so we keep the normalized OT-vs-CAL direction as the primary overthinking axis,
\(\hat{\mathbf{v}}_{\mathrm{OT}}^{(\ell)} \equiv
\hat{\mathbf{v}}_{\mathrm{OT}\text{-}\mathrm{vs}\text{-}\mathrm{CAL}}^{(\ell)}\),
and orthogonalize OA-vs-CAL against it:
\begin{equation}
  \hat{\mathbf{v}}_{\mathrm{OA}\perp}^{(\ell)}
=
\frac{
\hat{\mathbf{v}}_{\mathrm{OA}\text{-}\mathrm{vs}\text{-}\mathrm{CAL}}^{(\ell)}
-
\left(
\hat{\mathbf{v}}_{\mathrm{OA}\text{-}\mathrm{vs}\text{-}\mathrm{CAL}}^{(\ell)}
\cdot
\hat{\mathbf{v}}_{\mathrm{OT}}^{(\ell)}
\right)
\hat{\mathbf{v}}_{\mathrm{OT}}^{(\ell)}
}{
\left\|
\hat{\mathbf{v}}_{\mathrm{OA}\text{-}\mathrm{vs}\text{-}\mathrm{CAL}}^{(\ell)}
-
\left(
\hat{\mathbf{v}}_{\mathrm{OA}\text{-}\mathrm{vs}\text{-}\mathrm{CAL}}^{(\ell)}
\cdot
\hat{\mathbf{v}}_{\mathrm{OT}}^{(\ell)}
\right)
\hat{\mathbf{v}}_{\mathrm{OT}}^{(\ell)}
\right\|_2
}.
\label{eq:gram_schmidt}
\end{equation}
This decoupling ensures that steering along one axis does not perturb the projection onto the other.

We also estimate the calibrated centroid and projection scale from the axis-training split.
For each selected layer $\ell$, let
\begin{equation}
  \boldsymbol{\mu}_{\text{CAL}}^{(\ell)}
  =
  \frac{1}{|S_{\text{CAL}}|}
  \sum_{t \in S_{\text{CAL}}}
  \mathbf{h}_t^{(\ell)}
  \label{eq:cal_centroid}
\end{equation}
be the calibrated centroid: the empirical mean of hidden states for steps the judge labeled CAL, which serves as the reference point that steering pulls drifted activations back toward.
For each axis $a \in \{\text{OT},\text{OA}^\perp\}$, we compute the calibrated projection standard deviation
\begin{equation}
  \sigma_{\text{CAL},a}^{(\ell)}
  =
  \mathrm{std}_{t \in S_{\text{CAL}}}
  \left[
  \left(\mathbf{h}_t^{(\ell)}-\boldsymbol{\mu}_{\text{CAL}}^{(\ell)}\right)
  \cdot
  \hat{\mathbf{v}}_{a}^{(\ell)}
  \right].
  \label{eq:cal_sigma}
\end{equation}
This sets a per-axis scale for how far calibrated steps typically deviate along $\hat{\mathbf{v}}_{a}^{(\ell)}$, so that a threshold expressed in units of $\sigma_{\text{CAL},a}^{(\ell)}$ has a consistent meaning across layers.
Together, $\boldsymbol{\mu}_{\text{CAL}}^{(\ell)}$ and $\sigma_{\text{CAL},a}^{(\ell)}$ define the calibrated band used by test-time steering (Algorithm~\ref{alg:axis} in Appendix~\ref{app:algorithm}).

\subsection{Test-time Steering with Agent Drift Axes}
\label{sec:steering}

We implement steering as a hook on the residual stream that runs while the agent is rolling out, with no fine-tuning and no prompt changes.
Steering is the only thing that differs between the baseline rollouts and the steered rollouts in our experiments.

\textbf{Where it fires.} We register a forward hook at each of the top-$K$ transformer layers, ranked by validation per-problem AUC (Section~\ref{sec:experiments}).
The hook fires only on the forward pass that produces the \texttt{</think>} token of each step, where it reads the hidden state $\mathbf{h}_t^{(\ell)} \in \mathbb{R}^d$, modifies it along that layer's two drift axes, and writes the result back.

\begin{algorithm}[ht]
\caption{Dual-axis test-time steering (one forward pass at token $t$)}\label{alg:steer}
\KwInput{Per-layer axes $(\hat{\mathbf{v}}_{\text{OT}}^{(\ell)}, \hat{\mathbf{v}}_{\text{OA}^\perp}^{(\ell)})$, centroid $\boldsymbol{\mu}_{\text{CAL}}^{(\ell)}$, threshold $\tau_{a,\ell}$ for each $\ell \in \mathcal{L}$; strategy $\in \{\textsc{Add}, \textsc{Cap}, \textsc{Gated}\}$ with hyperparameters $(s, \lambda)$}
\For{$\ell \in \mathcal{L}$}{
  \uIf{strategy = \textsc{Add}}{
    $\mathbf{h}_t^{(\ell)} \leftarrow \mathbf{h}_t^{(\ell)} + s\,(\hat{\mathbf{v}}_{\text{OT}}^{(\ell)} + \hat{\mathbf{v}}_{\text{OA}^\perp}^{(\ell)})$
  }
  \Else(\Comment*[f]{\textsc{Cap}: $\beta=1$;\quad \textsc{Gated}: $\beta=\lambda$}){
    \For{$a \in \{\text{OT}, \text{OA}^\perp\}$}{
      $p \leftarrow (\mathbf{h}_t^{(\ell)}-\boldsymbol{\mu}_{\text{CAL}}^{(\ell)}) \cdot \hat{\mathbf{v}}_a^{(\ell)}$\;
      \lIf{$|p| > \tau_{a,\ell}$}{$\mathbf{h}_t^{(\ell)} \leftarrow \mathbf{h}_t^{(\ell)} - \beta\,\mathrm{sign}(p)\,(|p| - \tau_{a,\ell})\,\hat{\mathbf{v}}_a^{(\ell)}$}
    }
  }
}
\end{algorithm}

\textbf{What it modifies.} At every selected layer $\ell$, we steer along the orthogonalized pair from Section~\ref{sec:axis}: $\hat{\mathbf{v}}_{\text{OT}}^{(\ell)}$ (primary, from OT-vs-CAL) and $\hat{\mathbf{v}}_{\text{OA}^\perp}^{(\ell)}$ (secondary, orthogonalized from OA-vs-CAL).
For a given axis $a$ at layer $\ell$, the hook first computes the centered scalar projection
\begin{equation}
  p_{t,a}^{(\ell)}
  =
  \left(\mathbf{h}_t^{(\ell)}-\boldsymbol{\mu}_{\text{CAL}}^{(\ell)}\right)
  \cdot
  \hat{\mathbf{v}}_{a}^{(\ell)}
  \label{eq:projection}
\end{equation}
which says how far the current hidden state has drifted along that axis from the calibrated centroid.
Around the question of, given $p_{t,a}^{(\ell)}$, how much to subtract back along $\hat{\mathbf{v}}_{a}^{(\ell)}$, we design three test-time steering methods below (full procedure in Algorithm~\ref{alg:steer}).
The chosen method is applied independently to $\hat{\mathbf{v}}_{\text{OT}}^{(\ell)}$ and $\hat{\mathbf{v}}_{\text{OA}^\perp}^{(\ell)}$ at every selected layer.

\textbf{Method 1: Bidirectional Capping (\textsc{TACT-Cap}).} 
Bidirectional capping treats the calibrated region as a two-sided band along each drift axis.
For each axis \(a \in \{\mathrm{OT}, \mathrm{OA}^{\perp}\}\), it clamps the centered projection
\(p_{t,a}^{(\ell)}\) into \([-\tau_{a,\ell},+\tau_{a,\ell}]\), leaving in-band activations unchanged and pulling out-of-band activations back to the nearest band boundary:
\begin{equation}
  \mathbf{h}_t^{(\ell)} \leftarrow
    \begin{cases}
      \mathbf{h}_t^{(\ell)} - (p_{t,a}^{(\ell)} - \tau_{a,\ell})\,\hat{\mathbf{v}}_{a}^{(\ell)} & \text{if } p_{t,a}^{(\ell)} > \tau_{a,\ell}, \\
      \mathbf{h}_t^{(\ell)} - (p_{t,a}^{(\ell)} + \tau_{a,\ell})\,\hat{\mathbf{v}}_{a}^{(\ell)} & \text{if } p_{t,a}^{(\ell)} < -\tau_{a,\ell}, \\
      \mathbf{h}_t^{(\ell)} & \text{otherwise.}
    \end{cases}
  \label{eq:capping}
\end{equation}
The threshold is set as
\(\tau_{a,\ell}=k\cdot\sigma_{\mathrm{CAL},a}^{(\ell)}\), where
\(\sigma_{\mathrm{CAL},a}^{(\ell)}\) is the calibrated projection standard deviation and \(k\) controls the band width.
Although the positive direction of each axis corresponds to the named failure mode, unusually negative projections also fall outside the empirical calibrated distribution.
We therefore clip both tails, making capping conservative.

\textbf{Method 2: Additive (\textsc{TACT-Add}).}
Additive steering applies a fixed shift at every forward pass, regardless of the current projection.
At each selected layer, it moves the hidden state along the two steering directions with a scalar strength \(s\):
\begin{equation}
  \mathbf{h}_t^{(\ell)}
  \leftarrow
  \mathbf{h}_t^{(\ell)}
  +
  s \cdot
  \left(
  \hat{\mathbf{v}}_{\mathrm{OT}}^{(\ell)}
  +
  \hat{\mathbf{v}}_{\mathrm{OA}^{\perp}}^{(\ell)}
  \right).
  \label{eq:additive}
\end{equation}
In our convention, negative \(s\) pushes the activation away from both failure-mode directions.
Unlike capping, additive steering does not require estimating a threshold and does not check whether the activation is already calibrated.
This makes it simple and strong, but also less selective: if \(|s|\) is too large, the same uniform shift can perturb otherwise well-calibrated activations.

\textbf{Method 3: Projection-Gated (\textsc{TACT-Gate}).}
Projection-gated steering combines capping's selectivity with a tunable correction strength.
Like capping, it fires only when the centered projection leaves the calibrated band,
\(|p_{t,a}^{(\ell)}|>\tau_{a,\ell}\).
But it does not necessarily return the projection exactly to the band boundary; instead, it subtracts a \(\lambda\)-scaled amount of the out-of-band excess:
\begin{equation}
  \mathbf{h}_t^{(\ell)}
  \leftarrow
  \mathbf{h}_t^{(\ell)}
  -
  \lambda \cdot
  \mathrm{excess}\!\left(p_{t,a}^{(\ell)}, \tau_{a,\ell}\right)
  \hat{\mathbf{v}}_a^{(\ell)} ,
  \label{eq:projection_gated}
\end{equation}
where \(\mathrm{excess}(p,\tau)=p-\tau\) if \(p>\tau\),
\(\mathrm{excess}(p,\tau)=p+\tau\) if \(p<-\tau\), and \(0\) otherwise.
Here, \(\lambda\) controls the correction strength:
\(\lambda=1\) recovers bidirectional capping, \(\lambda<1\) leaves residual drift outside the band, and \(\lambda>1\) over-corrects past the boundary.
\section{Experiments}
\label{sec:experiments}

In this section, we organize the experiments around three questions.
\textbf{(Q1)} Does steering along the drift axes improve resolve rate?
We compare an unsteered baseline against three steering operators (\textsc{TACT-Cap}, \textsc{TACT-Add}, \textsc{TACT-Gate}) on SWE-bench Verified, Terminal-Bench 2.0, and CLAW-Eval in Section~\ref{sec:main_result}.
\textbf{(Q2)} Which design choices in TACT drive the performance gain?
We ablate axis choice, layer selection, and per-method hyperparameters of our method in Section~\ref{sec:ablations}.
\textbf{(Q3)} How well does the drift axis capture agent drift in hidden-state space? We probe the axis at the step level (per-problem AUC), trajectory level (drift-ratio correlations), and across alternative judge labels in Section~\ref{sec:axis_evidence}.
Beyond the core questions, we further check robustness of our method, including (1) hyperparameter sweep in Appendix~\ref{app:full_ablation} and (2) judge label reliability against humans and alternative judges in Appendix~\ref{app:judge_reliability}. Full setup details are in Appendix~\ref{app:setup_details}.

\textbf{Benchmarks.} We evaluate on three coding benchmarks that require long trajectories.
\textbf{SWE-bench Verified}~\citep{jimenez2024swebench} contains 500 real GitHub issues in Python repositories, where each instance requires agents to localize a bug in a large codebase and produce a correct patch; solving these tasks typically requires long trajectories (50--250+ steps), which often surface overthinking or overacting.
\textbf{Terminal-Bench 2.0}~\citep{tbench2025} contains 89 hard terminal tasks across 10 technical domains (bash, Python, system administration, environment setup, debugging), each run in a Docker sandbox with pytest-based verification; its diverse action space and shorter horizons expose overacting failures more often.
\textbf{CLAW-Eval}~\citep{claweval2026} contains 300 human-verified agent tasks spanning 9 categories, where agent performance is judged by rubric-based metrics and we report pass@3 as the resolve-rate analogue.
We use two splits: \textit{general} (single-agent service orchestration) and \textit{multi-turn} (user-agent dialogue where a second LLM plays the user), which together probe long-horizon reliability under both autonomous and interactive control.

\textbf{Baselines.} We compare \textsc{TACT} against four baselines from two paradigms that mitigate drift outside the model.
\textbf{Pre-failure prompt-level reminders} inject a fixed instruction every turn:
(i) \textsc{Reminder} (ours) appends a two-clause anti-drift instruction that mirrors the OT/OA definitions in Section~\ref{sec:formulation};
(ii) \textsc{ReCAP}~\citep{recap2025} re-injects the parent plan at every recursive return.
\textbf{Post-failure reflective feedback} runs a separate reflector after drift may have surfaced:
(iii) \textsc{AgentReflect} (ours) pauses every $K{=}10$ steps and prepends a reflector's three-line summary (progress, stuck signals, next direction) to the next turn;
(iv) \textsc{AgentDebug}~\citep{zhang2025agentdebug} root-causes a failed trajectory and feeds the corrective feedback into a retry.
\textsc{TACT} differs along two axes: it acts \emph{earlier}, modifying hidden states at the \texttt{</think>} token before any drift token is emitted, and \emph{more selectively}, firing only when the projection onto a drift axis exceeds the calibrated band $\tau_{a,\ell}$. Full experimental setup details are in Appendix~\ref{app:setup_details}, and baseline prompts and protocols are in Appendix~\ref{app:baselines}.

\subsection{Main Results}
\label{sec:main_result}

We evaluate the four external baselines and three \textsc{TACT} variants (\textsc{TACT-Cap}, \textsc{TACT-Add}, \textsc{TACT-Gate} in Section~\ref{sec:steering}) across Qwen3.5-27B and Gemma-4-26B-A4B-it.

\begin{table}[t]
  \caption{Main result across two open-weight reasoning models on SWE-bench Verified (SWE-V), Terminal-Bench 2.0 (TB2), and CLAW-Eval (CLAW).
  Res. refers to resolve rate; Steps refers to steps-to-resolve. \textsuperscript{$\dagger$}\textsc{AgentDebug} requires at least two rollouts per task (initial attempt + retry on failure); all other methods, including \textsc{TACT}, complete each task in a single rollout.}
  \label{tab:main}
  \centering
  \resizebox{\textwidth}{!}{%
  \begin{tabular}{llccccccc}
    \toprule
    & & \multicolumn{2}{c}{SWE-V} & \multicolumn{2}{c}{TB2} & \multicolumn{2}{c}{CLAW} & Avg.\ Res.\ \\
    \cmidrule(lr){3-4} \cmidrule(lr){5-6} \cmidrule(lr){7-8}
    Model & Method & Res.\ $\uparrow$ & Steps $\downarrow$ & Res.\ $\uparrow$ & Steps $\downarrow$ & Pass@3 \ $\uparrow$ & Steps $\downarrow$ & $\uparrow$ \\
    \midrule
    \multirow{8}{*}{Qwen3.5-27B}
      & Base Model             & 67.0 & 75.2 & 30.3 & 48.2 & 55.9 & 12.3 & 51.1 \\
      & \textsc{Reminder}        & 66.2 & 78.0 & 29.8 & 50.4 & 55.5 & 12.7 & 50.5 \\
      & \textsc{ReCAP}           & 67.9 & 75.8 & 31.1 & 49.5 & 56.4 & 12.5 & 51.8 \\
      & \textsc{AgentReflect}    & 68.4 & 73.1 & 32.0 & 46.8 & 57.0 & 11.9 & 52.5 \\
      & \textsc{AgentDebug}\textsuperscript{$\dagger$}      & 70.5 & 116.4 & 32.8 & 81.7 & 58.2 & 16.9 & 53.8 \\
    \cmidrule(lr){2-9}
      & \cellcolor{ysultralight}\textsc{TACT-Cap} \textit{(Ours)}  & \cellcolor{ysultralight}69.1 & \cellcolor{ysultralight}\textbf{55.7} & \cellcolor{ysultralight}32.6 & \cellcolor{ysultralight}\textbf{36.5} & \cellcolor{ysultralight}57.8 & \cellcolor{ysultralight}\textbf{9.4}  & \cellcolor{ysultralight}53.2 \\
      & \cellcolor{ysultralight}\textsc{TACT-Add} \textit{(Ours)}  & \cellcolor{ysultralight}73.0 & \cellcolor{ysultralight}67.5 & \cellcolor{ysultralight}35.9 & \cellcolor{ysultralight}42.7 & \cellcolor{ysultralight}61.2 & \cellcolor{ysultralight}11.1 & \cellcolor{ysultralight}56.7 \\
      & \cellcolor{ysultralight}\textsc{TACT-Gate} \textit{(Ours)} & \cellcolor{ysultralight}\textbf{73.3} & \cellcolor{ysultralight}65.3 & \cellcolor{ysultralight}\textbf{36.0} & \cellcolor{ysultralight}41.2 & \cellcolor{ysultralight}\textbf{61.5} & \cellcolor{ysultralight}10.8 & \cellcolor{ysultralight}\textbf{56.9} \\
    \midrule
    \multirow{8}{*}{Gemma-4-26B-A4B-it}
      & Base Model             & 59.7 & 52.1 & 15.7 & 18.2 & 56.3 & 9.2  & 43.9 \\
      & \textsc{Reminder}        & 60.0 & 53.2 & 15.9 & 18.6 & 56.8 & 9.3  & 44.2 \\
      & \textsc{ReCAP}           & 61.0 & 52.5 & 16.7 & 18.4 & 57.6 & 9.1  & 45.1 \\
      & \textsc{AgentReflect}    & 60.4 & 51.0 & 16.0 & 17.7 & 56.9 & 9.0  & 44.4 \\
      & \textsc{AgentDebug}\textsuperscript{$\dagger$}      & 62.0 & 78.9 & 17.2 & 31.5 & 58.0 & 12.6 & 45.7 \\
    \cmidrule(lr){2-9}
      & \cellcolor{ysultralight}\textsc{TACT-Cap} \textit{(Ours)}  & \cellcolor{ysultralight}61.6 & \cellcolor{ysultralight}\textbf{38.6} & \cellcolor{ysultralight}17.0 & \cellcolor{ysultralight}\textbf{13.9} & \cellcolor{ysultralight}58.2 & \cellcolor{ysultralight}\textbf{7.0} & \cellcolor{ysultralight}45.6 \\
      & \cellcolor{ysultralight}\textsc{TACT-Add} \textit{(Ours)}  & \cellcolor{ysultralight}65.2 & \cellcolor{ysultralight}46.2 & \cellcolor{ysultralight}18.9 & \cellcolor{ysultralight}16.5 & \cellcolor{ysultralight}61.0 & \cellcolor{ysultralight}8.4 & \cellcolor{ysultralight}48.4 \\
      & \cellcolor{ysultralight}\textsc{TACT-Gate} \textit{(Ours)} & \cellcolor{ysultralight}\textbf{65.5} & \cellcolor{ysultralight}44.7 & \cellcolor{ysultralight}\textbf{19.1} & \cellcolor{ysultralight}16.0 & \cellcolor{ysultralight}\textbf{61.4} & \cellcolor{ysultralight}8.1 & \cellcolor{ysultralight}\textbf{48.7} \\
    \bottomrule
  \end{tabular}%
  }
  \vspace{-3mm}
\end{table}

\textbf{Baselines.} As shown in Table~\ref{tab:main}, \textsc{Reminder} injects a direct per-turn prompt nudge against overthinking and overacting (Appendix~\ref{app:reminder_prompt}), but does not significantly improve resolve rate on either model ($51.1\% \to 50.5\%$ on Qwen, $43.9\% \to 44.2\%$ on Gemma). \textsc{ReCAP} and \textsc{AgentReflect} each add about $+1$\,pp on average. \textsc{AgentDebug} is the strongest baseline at $+2.7$\,pp on Qwen and $+1.8$\,pp on Gemma, but pays for the gain with at least two rollouts per task; the steps grow to $1.4$--$1.7{\times}$ compared to other methods for each rollout.

\textbf{\textsc{TACT}.} In addition, all three TACT variants improve resolve rate over all setups, while adding no extra LLM calls, prompt tokens, or retries; \textsc{TACT-Gate} gives the largest resolve gain ($+5.8$\,pp on Qwen, $+4.8$\,pp on Gemma), with \textsc{TACT-Add} close behind. \textsc{TACT-Cap} gives a smaller resolve gain but reduces steps-to-resolve the most, by $26\%$ on SWE-V for both models. \textsc{TACT-Cap} fires only when an activation leaves the calibrated band, so it reduces unnecessary steps without perturbing the rest of the trajectory; \textsc{TACT-Add} and \textsc{TACT-Gate} act on most steps with smaller shifts, which accumulate into a behavioral bias that lifts resolve rate. Compared with \textsc{AgentDebug}, the \textsc{TACT} activation hook is a constant-time read-write on the residual stream, which lets \textsc{TACT-Cap} match \textsc{AgentDebug}'s resolve gain at half the rollout cost ($+2.1$ vs $+2.7$\,pp on Qwen, $+1.7$ vs $+1.8$\,pp on Gemma), and \textsc{TACT-Gate} exceed it by roughly $+3$\,pp on both models.

\subsection{Ablations}
\label{sec:ablations}

We ablate two design choices in \textsc{TACT} as shown in Table~\ref{tab:ablations}: (i) which axes we steer along (dual vs.\ single, with vs.\ without orthogonalization), and (ii) which transformer layers we hook (top-$K$ by validation AUC vs.\ all). We ablate on \textsc{TACT-Gate} with fixed config $\lambda{=}0.5$ on SWE-bench Verified with Qwen3.5-27B; more detailed per-method hyperparameter sweeps are in Appendix~\ref{app:full_ablation}.

\begin{table}[t]
  \caption{Ablation on SWE-bench Verified with Qwen3.5-27B, \textsc{TACT-Gate} with $\lambda=0.5$. Each row removes one component of the full method. Columns under \textit{OT axis} and \textit{OA axis} mark whether each contrastive direction is used; \textit{Orthog} marks whether Gram-Schmidt is applied to make the two axes orthogonal; \textit{Layers} reports which transformer layers are hooked. We choose top-$K$ layers by validation per-problem AUC. $\Delta$ is the gain in resolve rate over the unsteered baseline.}
  \label{tab:ablations}
  \centering
  \small
  \begin{tabular*}{\linewidth}{@{\extracolsep{\fill}}lccccccc@{}}
    \toprule
    Variant & OT axis & OA axis & Orthog & Layers & Resolve (\%) $\uparrow$ & $\Delta$ & Steps $\downarrow$ \\
    \midrule
    Baseline (no steering) & --         & --         & --         & --     & 67.0 & --   & 75.2 \\
    \midrule
    Full method (main)     & \checkmark & \checkmark & \checkmark & top-10 & \textbf{73.3} & \textbf{+6.3} & \textbf{65.3} \\
    \quad w/o OT axis      & --         & \checkmark & --         & top-10 & 69.6 & +2.6 & 71.5 \\
    \quad w/o OA axis      & \checkmark & --         & --         & top-10 & 71.2 & +4.2 & 69.7 \\
    \quad w/o orthogonalization & \checkmark & \checkmark & --    & top-10 & 70.5 & +3.5 & 70.0 \\
    \quad w/o layer selection & \checkmark & \checkmark & \checkmark & all 64 & 68.4 & +1.4 & 73.9 \\
    \midrule
    Random orthogonal axes & --         & --         & \checkmark & top-10 & 66.4 & --0.6 & 77.1 \\
    Shuffled-label axes    & \checkmark & \checkmark & \checkmark & top-10 & 65.9 & --1.1 & 78.0 \\
    \bottomrule
  \end{tabular*}
  \vspace{-3mm}
\end{table}

\textbf{Axis choice.} We find dual orthogonalized axes outperform OT-only by $+2.1$\,pp and OA-only by $+3.7$\,pp on resolve rate, and outperform a non-orthogonalized dual by $+2.8$\,pp. It demonstrates overthinking and overacting often co-occur within a trajectory, steering along only one axis may leave the other failure mode uncorrected. The two failure modes are also not fully orthogonal in the residual stream, overthinking steps often carry some overacting signal and vice versa, so without Gram-Schmidt (Eq.~\ref{eq:gram_schmidt}), steering along one axis may perturb the projection along the other. Moreover, replacing the learned axes with norm-matched random orthogonal directions ($66.4\%$) or axes fit on shuffled task labels ($65.9\%$) slightly degrades performance below the base model, confirming that the gains come from the learned drift directions.

\textbf{Layer selection.} Hooking the top-10 layers ranked by validation AUC ($K{=}10$) outperforms hooking all 64 by $+4.9$\,pp on resolve rate ($73.3$ vs $68.4$) and cuts steps-to-resolve from $73.9$ to $65.3$. We select layers using OT-vs-CAL AUC because it provides a cleaner and stronger signal across layers (peaking at $0.91$ vs.\ $0.77$ for OA-vs-CAL in Figure~\ref{fig:AUC_and_lda}a), making it the natural primary criterion before orthogonalizing the secondary OA direction against it (Appendix~\ref{app:algorithm}). The drift signal is concentrated in the early-to-middle layers: OT-vs-CAL per-problem AUC peaks at $0.91$ on layer 3 and stays above $0.86$ through layer $\sim$20, then drops below $0.7$ in the late layers; hooking all 64 mixes this clean early signal with late-layer axes that are too noisy to steer with.

\subsection{Quality of Drift Axis}
\label{sec:axis_evidence}

\begin{figure}[t]
  \centering
  \begin{subfigure}[b]{0.49\linewidth}
    \includegraphics[width=\linewidth]{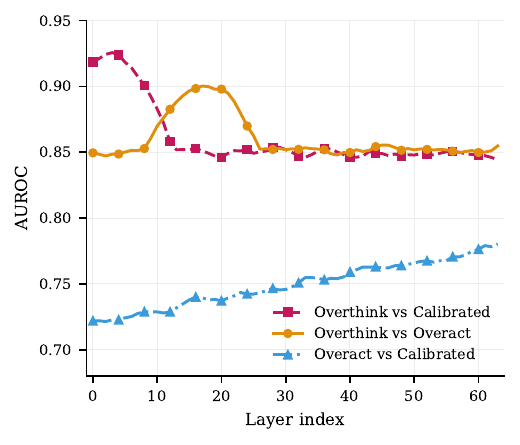}
    \caption{Per-problem AUC by layer}
  \end{subfigure}
  \hfill
  \begin{subfigure}[b]{0.49\linewidth}
    \includegraphics[width=\linewidth]{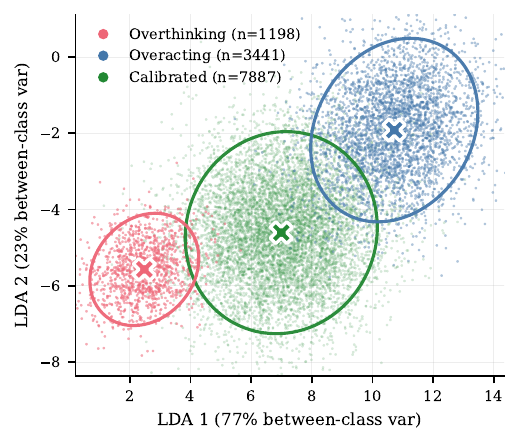}
    \caption{3-class LDA at layer 14}
  \end{subfigure}
  \caption{Quality of drift axes. (a) Per-problem AUC by transformer layer: OT-vs-CAL peaks at $0.91$ (layer 3), OT-vs-OA at $0.89$ (layer 17), OA-vs-CAL at $0.77$. (b) 3-class LDA projection at layer 14, showing OT/OA/CAL as separable clusters.}
  \label{fig:AUC_and_lda}
  \vspace{-4mm}
\end{figure}

\textbf{The axis carries drift signal.} We find OT-vs-CAL and OT-vs-OA AUCs both stay above $0.86$ through layer $\sim$20, peaking at $0.91$ and $0.89$, while OA-vs-CAL peaks at $0.77$ in later layers (Figure~\ref{fig:AUC_and_lda}a). It demonstrates drift is a stable, multi-layer feature in the residual stream rather than a single-layer artifact, so the contrastive axes pick up a property the model carries throughout its forward pass. As shown in Figure~\ref{fig:AUC_and_lda}b, the three classes are also separable in the 3-class LDA projection.

\textbf{What the drift axes encode.} To verify that the extracted axes capture genuine epistemic drift rather than surface heuristics, we conduct three interpretability and causal probes in Appendix~\ref{app:axis_interpretability}: (i) a manual audit of $300$ held-out steps shows that steps with extreme OT or OA projections match their target behavioral patterns in $86\%$ and $83\%$ of cases, respectively; (ii) regressing out reasoning token length and step index lowers per-problem AUC by only $0.04$ ($0.91 \to 0.87$ for OT-vs-CAL and $0.77 \to 0.73$ for OA-vs-CAL), ruling out simple length or position confounders; and (iii) steering along the positive OT or OA direction selectively inflates the corresponding failure mode without cross-contaminating the other. Additional analyses on frontier closed-source model cost comparisons, cross-benchmark axis transfer, and harmful flip rates are provided in Appendices~\ref{app:closed_models}--\ref{app:transfer_and_flips}.

\begin{table}[t]
  \caption{Judge-label robustness of the drift axis on SWE-bench Verified. Only the judge used to produce per-step labels differs; we keep the other configurations matching Table~\ref{tab:main}.}
  \label{tab:cross_judge_steer}
  \centering
  \small
  \begin{tabular*}{\linewidth}{@{\extracolsep{\fill}}llcc@{}}
    \toprule
    Label source & Method & Resolve (\%) $\uparrow$ & Steps $\downarrow$ \\
    \midrule
    --                   & Baseline           & 67.0 & 75.2  \\
    \midrule
    GPT-5.4 (main)       & \textsc{TACT-Cap}    & 69.1 & 55.7  \\
    GPT-5.4 (main)       & \textsc{TACT-Add}    & 73.0 & 67.5  \\
    GPT-5.4 (main)       & \textsc{TACT-Gate}   & 73.3 & 65.3  \\
    \midrule
    Claude Opus 4.6      & \textsc{TACT-Cap}    & 68.6 & 57.0  \\
    Claude Opus 4.6      & \textsc{TACT-Add}    & 72.5 & 68.4 \\
    Claude Opus 4.6      & \textsc{TACT-Gate}   & 72.7 & 66.6  \\
    \bottomrule
  \end{tabular*}
  \vspace{-6mm}
\end{table}

\textbf{Robustness to judge labels.} As shown in Table~\ref{tab:cross_judge_steer}, we find the end-to-end gain does not depend on which judge produced the per-step labels. We use Claude Opus~4.6 to rerun the labeling, and find that the resolve rate stays within $1$\,pp of GPT-5.4 for all three variants. Further reliability checks (human audit, cross-judge agreement) are in Appendix~\ref{app:judge_reliability}.


\section{Conclusion}
\label{sec:conclusion}

We presented TACT, a training-free pipeline that locates agent drift in the residual stream and corrects it at inference time. The pipeline labels each step as overthinking, overacting, or calibrated; extracts an orthogonalized pair of contrastive directions from these labels; and steers along the directions during rollouts. On SWE-bench Verified, Terminal-Bench 2.0, and CLAW-Eval, TACT raises average resolve rate by up to $+5.8$\,pp on Qwen3.5-27B and $+4.8$\,pp on Gemma-4-26B-A4B-it and cuts steps-to-resolve by up to $26\%$, while adding no extra LLM calls, prompt tokens, or retries. These gains frame agent drift as a steerable direction in the residual stream, and position TACT as a viable handle for reliable long-horizon agents. Future work can explore more complex activation interventions (e.g., sparse autoencoders) and extend the framework to non-coding domains.

\begin{ack}
\textbf{Funding:} The authors received no third-party funding or third-party support in direct support of this work during the last 36 months prior to submission.
\textbf{Competing Interests:} The authors declare no competing financial relationships or additional revenues outside the listed primary affiliations during the last 36 months that could be perceived to influence the submitted work.
\end{ack}

\bibliographystyle{plainnat}
\bibliography{references}

\clearpage
\appendix

\section*{Limitations}
\label{app:limitations}

\textbf{Domain scope.} \textsc{TACT} targets open-weight models where residual-stream activations are directly accessible, and our axes are fit and evaluated on software-engineering and tool-use benchmarks (SWE-bench, Terminal-Bench 2.0, CLAW-Eval). While agent drift is a broad phenomenon, the operational definitions of overthinking and overacting are domain-dependent. For instance, a web-browsing agent revisiting a page may be checking for live updates rather than looping, and a research agent re-querying with refined terms may be exploring rather than overacting. Extending \textsc{TACT} to non-coding domains therefore requires domain-appropriate judge rubrics rather than pooling trajectories across heterogeneous domains.

\textbf{Geometric vs.\ causal disentanglement and myopic intervention.} Gram-Schmidt orthogonalization (Eq.~\ref{eq:gram_schmidt}) guarantees that the OT and OA$^\perp$ updates do not directly alter each other's linear projection in the residual stream, but it is a geometric decoupling rather than a causal one. Furthermore, \textsc{TACT} operates greedily at the current step without lookahead: on genuinely difficult problems where an agent must explore dead-end hypotheses before finding a fix, genuine uncertainty can resemble overthinking in activation space, causing premature steering (accounting for the $\sim 3.2\%$ harmful flip rate analyzed in Appendix~\ref{app:transfer_and_flips}). Incorporating trajectory lookahead or outcome-conditioned judge labels, as well as nonlinear probes such as sparse autoencoders, are natural directions for future work.

\textbf{Reasoning formats and instruct-only models.} Our implementation hooks the \texttt{</think>} token because it provides an explicit reasoning-action boundary in reasoning models. Structured search patterns (e.g., MCTS or backtracking) remain compatible with our framework as long as new branches expand $\mathcal{K}_t$. For models with multi-stage tags (e.g., \texttt{<think>}, \texttt{<reflect>}, \texttt{<draft>}) or instruct-only models without explicit thinking tokens, the hook would need to be placed at the stage transition or immediately preceding the tool-call emission and refit on trajectories from that format. Finally, because \textsc{TACT} modifies only a single residual-stream state per step and leaves the sampling strategy and rollout budget untouched, it is orthogonal and complementary to test-time scaling methods such as best-of-$N$ sampling and budget forcing.

\section{Drift Axis Extraction}
\label{app:algorithm}

This section provides more details regarding the axis-extraction introduced in Section~\ref{sec:axis}.

\textbf{Pipeline at a glance.} Given a corpus of labeled trajectories, we (i) run one forward pass per labeled step and read the hidden state at the \texttt{</think>} token from every transformer layer; (ii) at each layer, compute two contrastive direction vectors, OT-vs-CAL and OA-vs-CAL, by per-problem balanced mean difference, then L2-normalize; (iii) at each layer, orthogonalize OA-vs-CAL against OT-vs-CAL via Gram-Schmidt to produce the dual pair $(\hat{\mathbf{v}}_{\text{OT}}^{(\ell)}, \hat{\mathbf{v}}_{\text{OA}^\perp}^{(\ell)})$; (iv) at each layer, estimate the calibrated centroid $\boldsymbol{\mu}_{\text{CAL}}^{(\ell)}$ and the per-axis projection standard deviation $\sigma_{\text{CAL},a}^{(\ell)}$ on calibrated training steps only; (v) on the held-out Axis-val split, rank layers by per-problem AUC of the OT-vs-CAL axis and keep the top-$K$ as the steering layer set $\mathcal{L}$. Algorithm~\ref{alg:axis} states the full procedure.

\begin{algorithm}[ht]
\caption{Drift axis extraction (per-layer)}\label{alg:axis}
\KwInput{Labeled steps $\{(\mathbf{h}_i^{(\ell)}, y_i)\}$ at \texttt{</think>} from $L$ layers; problem IDs $\{j_i\}$; calibrated step set $S_{\text{CAL}}$; problems-with-both-classes set $J$}
\KwOutput{For each $\ell$: orthogonalized pair $(\hat{\mathbf{v}}_{\text{OT}}^{(\ell)}, \hat{\mathbf{v}}_{\text{OA}^\perp}^{(\ell)})$, calibrated centroid $\boldsymbol{\mu}_{\text{CAL}}^{(\ell)}$, calibrated std $\sigma_{\text{CAL},a}^{(\ell)}$; selected layer set $\mathcal{L}$}
\BlankLine
\For{$\ell \in \{1, \dots, L\}$}{
  \tcp{Per-problem balanced mean-difference axes (Eq.~\ref{eq:axis})}
  \For{$(A, B) \in \{(\textsc{OT}, \textsc{CAL}), (\textsc{OA}, \textsc{CAL})\}$}{
    $\mathbf{v}_{A\text{-vs-}B}^{(\ell)} \leftarrow \frac{1}{|J|} \sum_{j \in J}\big( \overline{\mathbf{h}}_{A}^{j,(\ell)} - \overline{\mathbf{h}}_{B}^{j,(\ell)} \big)$ \;
    $\hat{\mathbf{v}}_{A\text{-}B}^{(\ell)} \leftarrow \mathbf{v}_{A\text{-vs-}B}^{(\ell)} / \|\mathbf{v}_{A\text{-vs-}B}^{(\ell)}\|_2$ \;
  }
  \tcp{Gram-Schmidt: keep OT-vs-CAL primary, orthogonalize OA-vs-CAL (Eq.~\ref{eq:gram_schmidt})}
  $\hat{\mathbf{v}}_{\text{OT}}^{(\ell)} \leftarrow \hat{\mathbf{v}}_{\text{OT-vs-CAL}}^{(\ell)}$ \;
  $\mathbf{u} \leftarrow \hat{\mathbf{v}}_{\text{OA-vs-CAL}}^{(\ell)} - \big(\hat{\mathbf{v}}_{\text{OA-vs-CAL}}^{(\ell)} \cdot \hat{\mathbf{v}}_{\text{OT}}^{(\ell)}\big)\, \hat{\mathbf{v}}_{\text{OT}}^{(\ell)}$ \;
  $\hat{\mathbf{v}}_{\text{OA}^\perp}^{(\ell)} \leftarrow \mathbf{u} / \|\mathbf{u}\|_2$ \;
  \tcp{Calibrated band statistics (Eqs.~\ref{eq:cal_centroid}, \ref{eq:cal_sigma})}
  $\boldsymbol{\mu}_{\text{CAL}}^{(\ell)} \leftarrow \frac{1}{|S_{\text{CAL}}|}\sum_{i \in S_{\text{CAL}}} \mathbf{h}_i^{(\ell)}$ \;
  \For{$a \in \{\text{OT}, \text{OA}^\perp\}$}{
    $\sigma_{\text{CAL},a}^{(\ell)} \leftarrow \mathrm{std}_{i \in S_{\text{CAL}}}\big[ (\mathbf{h}_i^{(\ell)} - \boldsymbol{\mu}_{\text{CAL}}^{(\ell)}) \cdot \hat{\mathbf{v}}_a^{(\ell)} \big]$ \;
  }
}
\BlankLine
\tcp{Layer selection on Axis-val split}
$\mathcal{L} \leftarrow$ top-$K$ layers by validation per-problem AUC of $\hat{\mathbf{v}}_{\text{OT-vs-CAL}}^{(\ell)}$ \;
\end{algorithm}

Regarding the design choices in the procedure, we have the following justifications:

(1) \textbf{Why OT-vs-CAL is the primary axis.} On Qwen3.5-27B, OT-vs-CAL reaches a per-problem AUC of $0.91$ at its best layer (layer 3) and stays above $0.86$ across most of the network, while OA-vs-CAL only reaches $0.77$ at its best layer and falls below $0.7$ across the early and middle layers (Figure~\ref{fig:AUC_and_lda}a). The 0.14-point gap at peak makes OT-vs-CAL the cleaner signal for layer selection and the more reliable direction to preserve under orthogonalization. We orthogonalize the weaker axis against the stronger one, so the weaker axis only loses its overlap with the cleaner direction. The 3-class LDA projection in Figure~\ref{fig:AUC_and_lda}b shows the same separation geometrically: the OT, OA, and CAL clusters are linearly separable at a representative middle layer.

(2) \textbf{Measuring drift from calibrated behavior.} To decide when to intervene, we need to know how far the current state has drifted from calibrated behavior. The centroid $\boldsymbol{\mu}_{\text{CAL}}^{(\ell)}$ gives the reference point, and the standard deviation $\sigma_{\text{CAL},a}^{(\ell)}$ gives the unit of measurement along each axis. The threshold in Algorithm~\ref{alg:steer} is then $\tau_{a,\ell} = k\,\sigma_{\text{CAL},a}^{(\ell)}$, which is just $k$ standard deviations of normal calibrated behavior at that specific layer and axis.

(3) \textbf{Layer selection.} We rank layers by per-problem AUC of the OT-vs-CAL axis on Axis-val and keep the top-$K$, with $K = 10$ as the default.

(4) \textbf{Cost.} Axis extraction is cheap. After one forward pass to cache $\mathbf{h}_t^{(\ell)}$ at every layer, the rest of the pipeline is closed-form linear algebra and finishes in seconds on a single GPU.

(5) \textbf{Why orthogonalize per-layer.} The residual stream changes from layer to layer, so the direction that captures ``overthinking'' at layer 5 is a different vector from the corresponding direction at layer 12. We extract OT-vs-CAL and OA-vs-CAL independently at every layer, then apply Gram-Schmidt within that layer to remove the OT-vs-CAL component from OA-vs-CAL. The orthogonalized pair lives in the layer-$\ell$ residual space, and because the two axes are orthogonal, the OT correction and the OA$^\perp$ correction at test time (Algorithm~\ref{alg:steer}) can be applied in any order without affecting the result.

\section{Robustness and Extended Analyses}
\label{app:robustness}

This section reports five robustness and extended analyses: a test-time steering hyperparameter sweep (Appendix~\ref{app:full_ablation}), a reliability check of the judge labels (Appendix~\ref{app:judge_reliability}), interpretability and causal probing of the drift axes (Appendix~\ref{app:axis_interpretability}), a performance and cost comparison with frontier closed-source models (Appendix~\ref{app:closed_models}), and cross-benchmark transfer and flip analyses (Appendix~\ref{app:transfer_and_flips}).

\begin{table}[t]
  \caption{Per-method strength sweeps on SWE-bench Verified with Qwen3.5-27B, complementing Table~\ref{tab:ablations}. All rows use per-layer dual-axis orthogonalized pairs at the top-10 layers selected by validation per-problem AUC. $\Delta$ is the gain in resolve rate over the unsteered baseline.}
  \label{tab:ablations_full}
  \centering
  \small
  \begin{tabular*}{\linewidth}{@{\extracolsep{\fill}}lccc@{}}
    \toprule
    Variant & Resolve (\%) $\uparrow$ & $\Delta$ & Steps $\downarrow$ \\
    \midrule
    Baseline (no steering)                       & 67.0 & --   & 75.2 \\
    \midrule
    \textsc{TACT-Cap} (main, $k = 2.0$)          & \textbf{69.1} & \textbf{+2.1} & \textbf{55.7} \\
    \quad $k = 1.0$                              & 67.1 & +0.1 & 59.8 \\
    \quad $k = 1.5$                              & 68.3 & +1.3 & 56.4 \\
    \quad $k = 3.0$                              & 68.0 & +1.0 & 66.0 \\
    \midrule
    \textsc{TACT-Add} (main, $s = -0.5$)         & \textbf{73.0} & \textbf{+6.0} & \textbf{67.5} \\
    \quad $s = -1.0$                             & 71.1 & +4.1 & 70.0 \\
    \quad $s = -2.0$                             & 64.8 & --2.2 & 82.2 \\
    \quad $s = +0.5$                             & 63.0 & --4.0 & 82.8 \\
    \midrule
    \textsc{TACT-Gate} (main, $\lambda = 0.5$)   & \textbf{73.3} & \textbf{+6.3} & 65.3 \\
    \quad $\lambda = 1.0$ ($k=2.0$)              & \multicolumn{3}{c}{\textit{equivalent to TACT-Cap ($k=2.0$); see row above}} \\
    \quad $\lambda = 1.5$                        & 71.6 & +4.6 & \textbf{63.1} \\
    \quad $\lambda = 2.0$                        & 69.0 & +2.0 & 65.5 \\
    \bottomrule
  \end{tabular*}
\end{table}

\begin{table}[t]
  \caption{Judge vs.\ human consensus on $n{=}300$ audited steps (balanced 100/100/100 across OT/OA/CAL). Overall Cohen's $\kappa$ is $0.78$.}
  \label{tab:human_verify}
  \centering
  \small
  \begin{tabular}{lcccc}
    \toprule
    Class & Precision & Recall & $F_1$ & OT$\leftrightarrow$OA confusion \\
    \midrule
    OT  & 0.87 & 0.84 & 0.85 & 5\% (OT$\to$OA) \\
    OA  & 0.88 & 0.82 & 0.85 & 6\% (OA$\to$OT) \\
    CAL & 0.79 & 0.87 & 0.83 & -- \\
    \bottomrule
  \end{tabular}
\end{table}

\subsection{Full ablation sweeps}
\label{app:full_ablation}

Each steering method has a single scalar that controls how strongly it pulls the hidden state back toward calibrated behavior at every selected layer. We refer to it as the \emph{strength hyperparameter}: $k$ (\textsc{TACT-Cap} band width in standard deviations), $s$ (\textsc{TACT-Add} shift along the axes), and $\lambda$ (\textsc{TACT-Gate} correction strength). Table~\ref{tab:ablations_full} varies this scalar for each method on SWE-bench Verified with Qwen3.5-27B, with all other settings fixed (top-10 layers, dual-axis pair). The goal is to see how each method responds as we turn its strength knob.

\textbf{Each method has a sweet spot.} Every method has a best strength setting. \textsc{TACT-Cap} works best at $k = 2.0$ (resolve rate $69.1$, $+2.1$). \textsc{TACT-Add} works best at $s = -0.5$ ($73.0$, $+6.0$). \textsc{TACT-Gate} works best at $\lambda = 0.5$ ($73.3$, $+6.3$). A setting too weak does not remove enough drift, and a setting too strong starts to pull calibrated steps off course, so resolve rate drops on either side.

\textbf{Steps shrink even when resolve rate is flat.} Even a weak setting saves compute. \textsc{TACT-Cap} at $k = 1.0$ gives the same resolve rate as the baseline ($67.1$ vs $67.0$), but average steps fall from $75.2$ to $59.8$. The intervention makes trajectories shorter even on problems the agent would solve anyway.

\textbf{\textsc{TACT-Add} is the most aggressive, \textsc{TACT-Gate} the most stable.} The three methods are not equally forgiving when the strength is set wrong. \textsc{TACT-Add} is the touchiest: at $s = -2.0$ it loses $2.2$ points to the baseline, and at $s = +0.5$ it loses $4.0$ points. \textsc{TACT-Gate}, by contrast, stays within $4.3$ points of its peak across $\lambda \in \{0.5, 1.5, 2.0\}$. If the same strength has to cover many tasks, \textsc{TACT-Gate} is the safer pick. We omit $\lambda{=}1$ from the sweep because Eq.~\ref{eq:projection_gated} reduces algebraically to bidirectional capping at that setting, so its row is exactly the \textsc{TACT-Cap} ($k=2.0$) row.

\subsection{Judge label reliability}
\label{app:judge_reliability}

Figure~\ref{fig:judge}(a,b) summarizes the two reliability checks behind our judge labels: (a) agreement with a human consensus and (b) cross-model consistency against three alternative judges. This subsection provides more details on the procedures and results.

\textbf{Human verification.}\label{app:human_verify} We selected a sample of $n{=}300$ labeled steps from the trajectories, balanced across the three labels (100 OT, 100 OA, 100 CAL) and across short ($N{<}50$) and long ($N{\geq}150$) trajectories so the audit covers both early and late drift. Human annotators with prior SWE-bench experience then label each step as OT, OA, or CAL, and we compare the judge against the human consensus using Cohen's $\kappa$, per-class precision and recall, and the OT$\leftrightarrow$OA confusion entries that would most directly contaminate the contrastive axes. 

In Table~\ref{tab:human_verify}, the judge agrees with humans at $\kappa = 0.78$ (substantial agreement) and all three classes have $F_1$ above $0.83$. The CAL row has the lowest precision ($0.79$): among the steps the judge predicts as CAL, about $21\%$ are labeled OT or OA by humans, so the CAL bucket absorbs some borderline drift steps as false positives. Recall on CAL is high ($0.87$), so genuine calibrated steps are rarely missed. The most consequential failure for our pipeline is OT$\leftrightarrow$OA confusion, since it would mix the two contrastive classes that define the axes; this confusion is at $5$--$6\%$, low enough that the axis direction is dominated by correctly labeled steps.

\textbf{Cross-model consistency.}\label{app:cross_judge} A natural concern about the pipeline is that the labels rely too heavily on one specific judge model (we use GPT-5.4 as the default judge): if GPT-5.4 has its own biases, the extracted axis would just encode those biases. To check this, we rerun the same three-stage pipeline (Section~\ref{sec:judge}) with three alternative language models of comparable capability, Claude Opus~4.6, Kimi K2.6, Grok-4.1-fast. We compare their per-step labels against GPT-5.4, and find that Cohen's $\kappa$ falls in the substantial-agreement range (Figure~\ref{fig:judge}b), showing consistent labels across judges.

\subsection{Interpretability and causal probing of the drift axes}
\label{app:axis_interpretability}

To understand what the learned drift axes encode beyond linear separability, we conduct three complementary analyses on held-out trajectories from SWE-bench Verified with Qwen3.5-27B.

\begin{table}[t]
  \caption{Manual behavioral categorization of $300$ held-out steps stratified by drift-axis projection ($100$ highest OT projection, $100$ highest OA projection, and $100$ closest to the calibrated centroid).}
  \label{tab:axis_interpretability}
  \centering
  \small
  \begin{tabular*}{\linewidth}{@{\extracolsep{\fill}}lp{0.58\linewidth}c@{}}
    \toprule
    Projection group & Dominant behavioral patterns & Matches mode \\
    \midrule
    High-OT ($n{=}100$)       & Re-derives a known conclusion ($39\%$) $\cdot$ cycles explored hypotheses ($27\%$) $\cdot$ deliberates over a tool-solvable question ($20\%$) & $86\%$ \\
    High-OA ($n{=}100$)       & Re-reads known files ($34\%$) $\cdot$ re-runs tests ($26\%$) $\cdot$ repeats searches ($15\%$) $\cdot$ acts without processing observation ($8\%$) & $83\%$ \\
    Near-centroid ($n{=}100$) & Productive information-state update (localization, edit, or test) & $79\%$ \\
    \bottomrule
  \end{tabular*}
\end{table}

\textbf{Behavioral breakdown of extreme projections.} We sample $300$ held-out steps: the $100$ with the highest projections along $\hat{\mathbf{v}}_{\text{OT}}^{(\ell)}$, the $100$ with the highest projections along $\hat{\mathbf{v}}_{\text{OA}^\perp}^{(\ell)}$, and the $100$ closest to the calibrated centroid $\boldsymbol{\mu}_{\text{CAL}}^{(\ell)}$. Human annotators review the surrounding trajectories and categorize each step into fine-grained behavioral sub-types. As shown in Table~\ref{tab:axis_interpretability}, $86\%$ of High-OT steps and $83\%$ of High-OA steps exhibit concrete instances of their respective failure definitions, while $79\%$ of near-centroid steps represent productive information-state updates.

\textbf{Controlling for reasoning length and trajectory position.} A potential concern is that the drift axes merely capture surface correlates such as the number of reasoning tokens in $r_t$ or the step index $t$ within a long trajectory. To test this, we fit an ordinary least-squares regression predicting each step's scalar projection $p_{t,a}^{(\ell)}$ from its reasoning token count and step index, subtract the predicted value to obtain residualized projections, and recompute per-problem AUC. Removing both surface features reduces peak AUC only slightly: from $0.91$ to $0.87$ for OT-vs-CAL and from $0.77$ to $0.73$ for OA-vs-CAL. Thus, token length and trajectory position account for only a minor fraction of the separation captured by the drift axes.

\textbf{Causal specificity via positive single-axis steering.} To verify that each axis causally controls its intended behavior, we run controlled interventions on held-out tasks where we steer \emph{positively} along only $\hat{\mathbf{v}}_{\text{OT}}^{(\ell)}$ or only $\hat{\mathbf{v}}_{\text{OA}^\perp}^{(\ell)}$, and re-annotate the resulting rollouts with the judge pipeline. Steering along $+\hat{\mathbf{v}}_{\text{OT}}^{(\ell)}$ increases the trajectory OT ratio from $8.6\%$ to $22.3\%$ while changing the OA ratio by less than $2$\,pp; conversely, steering along $+\hat{\mathbf{v}}_{\text{OA}^\perp}^{(\ell)}$ selectively inflates OA steps while leaving OT nearly unchanged. This confirms that the two orthogonalized axes act as selective causal levers.

\subsection{Comparison with frontier closed-source models and cost analysis}
\label{app:closed_models}

Because fitting the drift axes uses an offline LLM-as-judge, a natural practical question is how \textsc{TACT}-steered open-weight models compare in performance and inference cost against frontier proprietary models run directly as the coding agent. Table~\ref{tab:closed_comparison} compares Qwen3.5-27B and Gemma-4-26B-A4B-it (with and without \textsc{TACT-Gate}) against Claude Opus~4.6, GPT-5.2-Codex, and Gemini~3~Pro under the exact same \textit{mini-swe-agent} scaffold on SWE-bench Verified. For self-hosted open-weight models, we estimate per-task inference cost using standard third-party API token pricing.

\begin{table}[t]
  \caption{Performance and per-task inference cost comparison on SWE-bench Verified under the same \textit{mini-swe-agent} scaffold. Open-weight costs are estimated from standard third-party API pricing.}
  \label{tab:closed_comparison}
  \centering
  \small
  \begin{tabular*}{\linewidth}{@{\extracolsep{\fill}}lccc@{}}
    \toprule
    Model / Method & SWE-V Resolve (\%) $\uparrow$ & Cost / Task ($\$$) $\downarrow$ & Relative to Opus 4.6 \\
    \midrule
    Claude Opus 4.6                        & 75.6 & $\approx 0.770$ & $1.0\times$ \\
    GPT-5.2-Codex                          & 72.8 & $\approx 0.630$ & $1.2\times$ cheaper \\
    Gemini 3 Pro                           & 69.6 & $\approx 1.340$ & $1.7\times$ more expensive \\
    \midrule
    Qwen3.5-27B (Base Model)               & 67.0 & $\approx 0.112$ & $6.9\times$ cheaper \\
    \cellcolor{ysultralight}\quad + \textsc{TACT-Gate} \textit{(Ours)} & \cellcolor{ysultralight}\textbf{73.3} & \cellcolor{ysultralight}$\approx \mathbf{0.097}$ & \cellcolor{ysultralight}$\mathbf{7.9\times}$ \textbf{cheaper} \\
    \midrule
    Gemma-4-26B-A4B-it (Base Model)        & 59.7 & $\approx 0.031$ & $24.8\times$ cheaper \\
    \cellcolor{ysultralight}\quad + \textsc{TACT-Gate} \textit{(Ours)} & \cellcolor{ysultralight}\textbf{65.5} & \cellcolor{ysultralight}$\approx \mathbf{0.026}$ & \cellcolor{ysultralight}$\mathbf{29.1\times}$ \textbf{cheaper} \\
    \bottomrule
  \end{tabular*}
\end{table}

\textsc{TACT-Gate} lifts Qwen3.5-27B from $67.0\%$ to $73.3\%$, surpassing both GPT-5.2-Codex ($72.8\%$) and Gemini~3~Pro ($69.6\%$) while costing roughly $7.9\times$ less per task than Claude Opus~4.6 ($\approx \$0.097$ vs.\ $\$0.77$). Moreover, because \textsc{TACT-Gate} shortens trajectories (from $75.2$ to $65.3$ steps on Qwen and from $52.1$ to $44.7$ steps on Gemma), it simultaneously reduces per-task inference cost relative to the unsteered base model. For deployments with strict privacy or self-hosting requirements, the offline judge can also be replaced by an open-weight model such as Kimi~K2.6 ($\kappa = 0.76$ agreement with GPT-5.4 in Figure~\ref{fig:judge}b), keeping the entire pipeline self-hosted.

\subsection{Cross-benchmark axis transfer and harmful flip analysis}
\label{app:transfer_and_flips}

\begin{table}[t]
  \caption{Zero-shot cross-benchmark transfer of drift axes extracted on SWE-bench Verified (Qwen3.5-27B) and applied directly to SWE-bench Multilingual and SWE-bench Pro without refitting.}
  \label{tab:cross_transfer}
  \centering
  \small
  \begin{tabular*}{\linewidth}{@{\extracolsep{\fill}}llccc@{}}
    \toprule
    Target Benchmark & Method & Resolve (\%) $\uparrow$ & $\Delta$ & Axis Source \\
    \midrule
    \multirow{2}{*}{SWE-bench Multilingual}
      & Base Model                      & 66.3 & --   & -- \\
      & \textsc{TACT-Gate} \textit{(Ours)} & \textbf{70.5} & \textbf{+4.2} & SWE-bench Verified (zero-shot) \\
    \midrule
    \multirow{2}{*}{SWE-bench Pro}
      & Base Model                      & 47.8 & --   & -- \\
      & \textsc{TACT-Gate} \textit{(Ours)} & \textbf{51.4} & \textbf{+3.6} & SWE-bench Verified (zero-shot) \\
    \bottomrule
  \end{tabular*}
\end{table}

\textbf{Cross-benchmark transfer.} To test whether the extracted drift axes generalize beyond the benchmark on which they were fit, we take the orthogonalized axes $(\hat{\mathbf{v}}_{\text{OT}}^{(\ell)}, \hat{\mathbf{v}}_{\text{OA}^\perp}^{(\ell)})$ and calibration statistics $(\boldsymbol{\mu}_{\text{CAL}}^{(\ell)}, \sigma_{\text{CAL},a}^{(\ell)})$ fit on SWE-bench Verified with Qwen3.5-27B and apply them zero-shot to \textbf{SWE-bench Multilingual} and \textbf{SWE-bench Pro} without re-annotation or hyperparameter retuning. As shown in Table~\ref{tab:cross_transfer}, \textsc{TACT-Gate} improves resolve rate from $66.3\%$ to $70.5\%$ ($+4.2$\,pp) on SWE-bench Multilingual and from $47.8\%$ to $51.4\%$ ($+3.6$\,pp) on SWE-bench Pro, demonstrating that the drift directions capture reusable representations across coding benchmarks.

\textbf{Harmful flip analysis.} While \textsc{TACT-Gate} intervenes only when a step's activation leaves the calibrated band $[-\tau_{a,\ell}, +\tau_{a,\ell}]$, steering is not completely risk-free: when an agent encounters a genuinely hard bug with insufficient evidence, legitimate deliberation can project positively onto the OT axis and trigger an early intervention. To quantify how often steering hurts a trajectory that would have succeeded unsteered, we randomly sample $100$ tasks from SWE-bench Verified and run both the unsteered Qwen3.5-27B model and \textsc{TACT-Gate} four times per task ($400$ paired rollouts). Defining a \emph{harmful flip} as a task where the unsteered rollout resolves the issue but the paired steered rollout fails (averaged across the four runs to control for sampling variance), we observe a harmful flip rate of approximately $3.2\%$, far outweighed by the positive flips that drive the net $+6.3$\,pp resolve gain.

\section{Experiment Setup}
\label{app:setup_details}

This section provides the remaining experimental setup details, complementing the benchmark and baseline descriptions in Section~\ref{sec:experiments}.

\textbf{Agent and models.} We collect trajectories on SWE-bench Verified using mini-swe-agent, a lightweight scaffold in which every action is a single bash tool call. This yields a clean boundary between $E_{\text{int}}$ (tokens inside \texttt{<think>...</think>}) and $E_{\text{ext}}$ (tool calls and observations). On Terminal-Bench 2.0 we wrap the same scaffold in its Dockerized harness; on CLAW-Eval we adopt the upstream harness with its native trace format (full details in Appendix~\ref{app:agent}).
Since hidden-state extraction and activation steering both require model internals, we study open-weight reasoning models whose native \texttt{<think>} blocks make the reasoning-action boundary explicit. We consider two models with different architectures: \textbf{Qwen3.5-27B} (dense, 64 layers, hidden dimension 5120, bfloat16) and \textbf{Gemma-4-26B-A4B-it}~\citep{gemma4}, a sparse mixture-of-experts with 26B total and $\sim$4B active parameters. Spanning dense versus MoE and different pretraining families lets us test whether the drift axis is a general property of reasoning-model activations rather than a Qwen-specific artifact.
We run the three-stage LLM-as-judge pipeline (Section~\ref{sec:judge}) with \textbf{GPT-5.4} as default. To verify label reliability, we conduct (i) a human audit and (ii) a cross-judge consistency check against Claude Opus~4.6, Kimi K2.6, and Grok-4.1-fast in Appendix~\ref{app:judge_reliability}; both show substantial agreement with the main judge.

\textbf{Labels and hidden states.} Each collected trajectory is labeled by the three-stage LLM-as-judge pipeline (Section~\ref{sec:judge}), producing a label (\textsc{overthinking}, \textsc{overacting}, or \textsc{calibrated}), a one-sentence justification, and a confidence level for every step.
Labeling the full axis-fitting corpus ($\sim 2\text{K}$ trajectories across models and benchmarks) required approximately $23\text{K}$ judge calls and $209\text{M}$ tokens at a one-time offline cost of roughly $\$600$; once the axes are extracted, test-time evaluation and deployment require zero judge calls.
We then run a single forward pass over the full conversation and extract the hidden state at the \texttt{</think>} token from every transformer layer.
All data splits are at the problem level, so no steps from the same task appear in more than one split.
We use three disjoint splits.
\textit{Axis-train} fits the mean-difference axes, calibrated centroids, and calibrated projection scales.
\textit{Axis-val} is used for model selection: we rank layers by per-problem AUC on the primary OT-vs-CAL axis, choose the top-$K$ layers, and select steering hyperparameters by validation resolve rate.
\textit{Eval} is reserved for the final reported resolve rate and steps, and is never used for fitting or tuning.

\textbf{Steering configurations.} All steering runs operate on per-layer orthogonalized axis pairs $\{(\hat{\mathbf{v}}_{\text{OT}}^{(\ell)}, \hat{\mathbf{v}}_{\text{OA}^\perp}^{(\ell)})\}_{\ell}$ (Section~\ref{sec:steering}), which correct both failure modes simultaneously within each selected layer since the axes are Euclidean-orthogonal and their interventions commute.
For each of the three methods (\textsc{TACT-Cap}, \textsc{TACT-Add}, \textsc{TACT-Gate}), we select the single best hyperparameter configuration on Axis-val by resolve rate, then apply that fixed configuration once to the evaluation set.
Hyperparameter sweeps: \textsc{TACT-Cap} $k \in \{1.0, 1.5, 2.0, 3.0\}$; \textsc{TACT-Add} $s \in \{-2.0, -1.0, -0.5, +0.5\}$; \textsc{TACT-Gate} $\lambda \in \{0.5, 1.5, 2.0\}$ at $k = 2.0$ ($\lambda{=}1$ is excluded because it reduces to \textsc{TACT-Cap} at the same $k$).
Unless noted, we steer the top-10 layers ranked by validation per-problem AUC on the primary axis.

\textbf{Metrics.} For each benchmark we report two metrics per instance, averaged over the evaluation set: \textit{resolve rate} (the primary success measure, defined per benchmark below) and \textit{average steps} (number of agent turns before submission or max-step cutoff).
For SWE-bench Verified and Terminal-Bench 2.0, resolve rate is the fraction of instances passing the benchmark's grading script.
CLAW-Eval does not expose a single pass/fail signal, so we use \textit{pass@3} (whether the rubric Completion grader passes on at least one of three independent trials) as the resolve-rate analogue.
Steps describe efficiency, since a good intervention should raise resolve rate without inflating trajectory length.

\section{Related Work}
\label{sec:related_work}

\textbf{Long-horizon LLM agents.}
LLM agents now solve tasks that span hundreds of steps: editing real GitHub repositories \citep{jimenez2024swebench, deng2025swebenchpro}, completing terminal workflows \citep{tbench2025}, navigating realistic web environments \citep{zhou2024webarena, mialon2024gaia}, controlling full operating systems \citep{xie2024osworld}, and exploring open-ended worlds \citep{wang2023voyager}.
A standard recipe combines tool use with structured reasoning: ReAct interleaves thoughts and actions \citep{yao2023react}, CodeAct executes code as the action space \citep{wang2024codeact}, and harnesses such as SWE-agent, OpenHands, and Agentless wrap a reasoning model with file-editing and shell tools \citep{yang2024sweagent, wang2024openhands, xia2024agentless, xia2025liveswe}.
Surveys catalog the resulting design space and the rapid pace of benchmark saturation \citep{wang2024agentsurvey, yehudai2025agenteval, guo2025seagentsurvey}.
Most of this work treats the agent as a black box and reports outcome metrics; we instead ask why long trajectories fail and locate the answer in the model's internal state.

\textbf{Agent drift, overthinking, and overacting.}
A growing body of work characterizes how language model agents degrade over long trajectories.
\citet{cuadron2025overthinking} show that reasoning-heavy models over-deliberate at the expense of external grounding; \citet{wang2025toa} argue the opposite symptom, that agents over-invoke tools when reasoning would suffice; and \citet{rath2026agentdrift} quantifies cumulative behavioral degradation under the name ``agent drift.''
Related planning analyses \citep{wang2026planning} find that reasoning and acting lose alignment as trajectories grow.
A parallel literature on reasoning models reports the same pathology in a single forward pass: optimal allocation of test-time compute is highly prompt-dependent \citep{snell2024testtimescaling, muennighoff2025s1}, thinking budgets that help small problems hurt large ones \citep{ghosal2025mirage}, and R1-style models construct redundant chains that survey work identifies as the dominant efficiency bottleneck \citep{yue2025dontoverthink, deepseekai2025r1, kimi2025k15}.
Long-context studies report a related ``lost in the middle'' effect that compounds with reasoning failures over long agent contexts \citep{liu2024lostmiddle}.
These studies converge on the diagnosis (agents behave worse as context accumulates) but stop at external metrics such as resolve rate or token count.
What is missing is a mechanistic handle: where in the model do overthinking and overacting live, and can we intervene on them directly?
We answer both questions by locating drift in the hidden-state geometry at the reasoning-action boundary.

\textbf{Self-improvement and inference-time control of reasoning.}
A separate thread tries to fix drift through verbal feedback or learned guidance rather than activation-level intervention.
Reflexion and Self-Refine ask the model to critique its own outputs and revise \citep{shinn2023reflexion, madaan2023selfrefine}; Tree of Thoughts and follow-ups search over alternative reasoning paths \citep{yao2023tot}; Agent-R trains agents to reflect via iterative self-training over recovered trajectories \citep{yuan2025agentr}.
Closer to the inference-time control we study, \citet{sui2025metareasoner} introduce a meta-reasoner that monitors a reasoning chain and dispatches strategies (backtrack, restart, switch approach) through a contextual bandit, and \citet{sui2026conl} use multi-agent meta-evaluation to provide reward signal in non-verifiable settings where ground-truth rewards do not exist.
\citet{sui2025whatif} train an LLM to act as a proactive world model that simulates how the environment will evolve under candidate actions before committing.
All three approaches keep the model fixed and intervene at the prompt or trajectory level.
We share their goal of steering reasoning behavior at inference time but operate one level lower, by editing hidden states along a learned direction at every reasoning-action boundary.

\textbf{Representation engineering and linear concepts.}
Many high-level behavioral properties are linearly encoded in transformer hidden states and can be steered with simple test-time interventions \citep{zou2023repeng, turner2023steering, panickssery2024contrastive, li2023iti, lee2024cast}.
The linear representation hypothesis formalizes this picture, identifying directions in activation space that correspond to interpretable concepts and admit causal interventions \citep{park2024linear, marks2024geometryoftruth}.
Sparse autoencoders extend the toolkit by decomposing activations into thousands of monosemantic features that can be ablated or amplified individually \citep{templeton2024scaling, gao2024sae, shu2025saesurvey, wang2025saerefine}.
Closest to our setting, \citet{lu2026assistant} study persona drift in Anthropic's assistant models and show that displacing activations back toward a stable assistant subspace stabilizes behavior over long contexts.
We share the same diagnosis-and-intervention recipe (identify a linear direction tied to a behavioral property, then steer along it) but target a different kind of drift: the think-act allocation of a tool-using agent over hundreds of steps.
Prior representation-engineering work focuses on single-turn outputs such as honesty, refusal, and persona; we extend the toolkit to behavioral failures that unfold across long agent trajectories, treating overthinking and overacting as linearly separable directions at the \texttt{</think>} token and steering them at test time during live multi-step agent rollouts.

\section{Agent Configuration}
\label{app:agent}

The three benchmarks are run under two different agent scaffolds.
\textbf{SWE-bench Verified} and \textbf{Terminal-Bench 2.0} both use \textit{mini-swe-agent}, a lightweight scaffold in which every environment interaction is routed through a single \texttt{bash} tool, so each step is either a reasoning block or one bash command. The two benchmarks share the same scaffold, model adapter, and observation template; only the system / instance templates and the submission convention differ. On SWE-bench Verified the agent runs against a per-instance repository checkout and submits a git patch; on Terminal-Bench 2.0 the agent runs inside a Harbor-provisioned Docker container and submits by issuing a sentinel command that the harness intercepts.
\textbf{CLAW-Eval} uses its upstream OpenClaw scaffold~\citep{claweval2026}. Unlike mini-swe-agent, OpenClaw exposes a per-task set of first-class tools (CRM, scheduler, web, OCR, knowledge base, document store, $\dots$) declared in each task's \texttt{task.yaml}; the agent's system prompt is composed at task load time from a static intro, the task's tool list, full JSON schemas, behavior rules, available skills, and workspace files (\texttt{AGENTS.md}, \texttt{SOUL.md}, \texttt{USER.md}, \texttt{TOOLS.md}). The \textit{multi-turn} split additionally runs a separate LLM as a simulated user with its own system prompt.

Subsections~\ref{app:tool}--\ref{app:observation} describe the SWE-bench Verified configuration of mini-swe-agent. Subsection~\ref{app:agent_tb2} describes the Terminal-Bench 2.0 deltas. Subsections~\ref{app:agent_claw_system}--\ref{app:agent_claw_user} describe the OpenClaw scaffold used on CLAW-Eval.

\subsection{Tool Definition}
\label{app:tool}

The agent has access to exactly one tool:

\begin{Verbatim}[frame=single,rulecolor=\color{ysdarkblue},fillcolor=\color{ysshallowblue},label=\textbf{Bash Tool (OpenAI Function Calling Schema)},fontsize=\small]
{
  "type": "function",
  "function": {
    "name": "bash",
    "description": "Execute a bash command",
    "parameters": {
      "type": "object",
      "properties": {
        "command": {
          "type": "string",
          "description": "The bash command to execute"
        }
      },
      "required": ["command"]
    }
  }
}
\end{Verbatim}

\subsection{Agent System Prompt}
\label{app:agent_system}

\begin{Verbatim}[frame=single,rulecolor=\color{ysdarkblue},fillcolor=\color{ysshallowblue},label=\textbf{System Prompt},fontsize=\small]
You are a helpful assistant that can interact with a computer
shell to solve programming tasks.
\end{Verbatim}

\subsection{Agent Instance Prompt}
\label{app:agent_instance}

\begin{Verbatim}[frame=single,rulecolor=\color{ysdarkblue},fillcolor=\color{ysshallowblue},label=\textbf{Instance Prompt (sent as the first user message)},fontsize=\small]
<pr_description>
Consider the following PR description:
{task}
</pr_description>

<instructions>
# Task Instructions

## Overview

You're a software engineer interacting continuously with a
computer by submitting commands. You'll be helping implement
necessary changes to meet requirements in the PR description.
Your task is specifically to make changes to non-test files in
the current directory in order to fix the issue described in
the PR description in a way that is general and consistent
with the codebase.
<IMPORTANT>This is an interactive process where you will think
and issue AT LEAST ONE command, see the result, then think and
issue your next command(s).</important>

For each response:

1. Include a THOUGHT section explaining your reasoning and
   what you're trying to accomplish
2. Provide one or more bash tool calls to execute

## Important Boundaries

- MODIFY: Regular source code files in /testbed
- DO NOT MODIFY: Tests, configuration files
  (pyproject.toml, setup.cfg, etc.)

## Recommended Workflow

1. Analyze the codebase by finding and reading relevant files
2. Create a script to reproduce the issue
3. Edit the source code to resolve the issue
4. Verify your fix works by running your script again
5. Test edge cases to ensure your fix is robust

## Command Execution Rules

You are operating in an environment where

1. You issue at least one command
2. The system executes the command(s) in a subshell
3. You see the result(s)
4. You write your next command(s)

Each response should include:

1. **Reasoning text** where you explain your analysis and plan
2. At least one tool call with your command

**CRITICAL REQUIREMENTS:**

- Your response SHOULD include reasoning text explaining
  what you're doing
- Your response MUST include AT LEAST ONE bash tool call.
  You can make MULTIPLE tool calls in a single response
  when the commands are independent.
- Directory or environment variable changes are not
  persistent. Every action is executed in a new subshell.

## Submission

When you've completed your work, you MUST submit your changes
as a git patch. Follow these steps IN ORDER, with SEPARATE
commands:

Step 1: Create the patch file
Run `git diff -- path/to/file1 path/to/file2 > patch.txt`

Step 2: Verify your patch

Step 3: Submit (EXACT command required)
You MUST use this EXACT command to submit:
echo COMPLETE_TASK_AND_SUBMIT_FINAL_OUTPUT && cat patch.txt
</instructions>
\end{Verbatim}

\subsection{Observation Template}
\label{app:observation}

Agent observations are rendered from tool outputs via a Jinja2 template.
Outputs exceeding 10{,}000 characters are truncated, keeping the first and last 5{,}000 characters:

\begin{Verbatim}[frame=single,rulecolor=\color{ysdarkblue},fillcolor=\color{ysshallowblue},label=\textbf{Observation Template (Jinja2)},fontsize=\small]
{% if output.exception_info -%}
<exception>{{output.exception_info}}</exception>
{% endif -%}
<returncode>{{output.returncode}}</returncode>
{% if output.output | length < 10000 -%}
<output>
{{ output.output -}}
</output>
{%- else -%}
<warning>
The output of your last command was too long.
Please try a different command that produces less output.
</warning>
<output_head>
{{ output.output[:5000] }}
</output_head>
<elided_chars>
{{ elided_chars }} characters elided
</elided_chars>
<output_tail>
{{ output.output[-5000:] }}
</output_tail>
{%- endif -%}
\end{Verbatim}

\subsection{Terminal-Bench 2.0: mini-swe-agent reuse}
\label{app:agent_tb2}

For Terminal-Bench 2.0 we wrap the same mini-swe-agent in the Harbor harness, which provisions one Docker container per task and installs the scaffold inside. The bash tool definition (Section~\ref{app:tool}) and observation template (Section~\ref{app:observation}) are unchanged. Only the system / instance templates and the submission convention differ: there is no PR description and the agent submits by issuing a sentinel command rather than a git patch.

\begin{Verbatim}[frame=single,rulecolor=\color{ysdarkblue},fillcolor=\color{ysshallowblue},label=\textbf{TB2 System Prompt},fontsize=\small]
You are a helpful assistant that can interact with a Linux
shell to solve terminal-based tasks.
\end{Verbatim}

\begin{Verbatim}[frame=single,rulecolor=\color{ysdarkblue},fillcolor=\color{ysshallowblue},label=\textbf{TB2 Instance Prompt (excerpt)},fontsize=\small]
<task>
{task}
</task>

<instructions>
You are working inside a Linux container. Your job is to
complete the task above by issuing shell commands. The task
will be evaluated by automated tests that inspect the state of
the container's filesystem, any produced output files, or the
behavior of commands you create.

For each response:
1. Include a THOUGHT section explaining your reasoning.
2. Provide one or more bash tool calls to execute.

CRITICAL REQUIREMENTS:
- Every response MUST include at least one bash tool call.
- Directory or environment variable changes are NOT persistent
  between turns; chain related work with `&&` or re-establish
  state at the start of each command.
- Use non-interactive flags. Avoid interactive tools like
  vi/nano; use `cat >`, `sed`, `awk`, `tee`, or heredocs.
- If a required tool is missing, install it.

When the task is complete, submit with this EXACT command:

    echo COMPLETE_TASK_AND_SUBMIT_FINAL_OUTPUT

If that command is not issued, the harness keeps prompting.
After submission you cannot continue working on this task.
</instructions>
\end{Verbatim}

\subsection{CLAW-Eval: OpenClaw scaffold}
\label{app:agent_claw_system}

OpenClaw exposes a richer per-task action space than mini-swe-agent. Each CLAW task ships with a \texttt{task.yaml} that declares the tools the agent may call (drawn from a fixed registry of mock services: CRM, scheduler, web, OCR, knowledge base, document store, contacts, notes, $\dots$). At task load time the scaffold composes a system prompt by concatenating six blocks: (i) a static intro identifying the harness; (ii) a tool inventory rendered from the task's tool list; (iii) full JSON schemas of those tools; (iv) behavior rules covering tool-call style and safety; (v) a skills block listing optional \texttt{SKILL.md} files the agent may load; and (vi) workspace files (\texttt{AGENTS.md}, \texttt{SOUL.md}, \texttt{USER.md}, \texttt{TOOLS.md}) injected verbatim. The first block is fixed across all tasks; the rest are task-conditional. The static parts are below.

\begin{Verbatim}[frame=single,rulecolor=\color{ysdarkblue},fillcolor=\color{ysshallowblue},label=\textbf{OpenClaw System Prompt -- Static Intro},fontsize=\small]
You are a personal assistant running inside OpenClaw.
\end{Verbatim}

\begin{Verbatim}[frame=single,rulecolor=\color{ysdarkblue},fillcolor=\color{ysshallowblue},label=\textbf{OpenClaw System Prompt -- Tool Inventory and Schemas (template)},fontsize=\small]
## Tooling
Tool availability (filtered by policy):
Tool names are case-sensitive. Call tools exactly as listed.
- {tool_name}: {tool_description}
- ...
When a first-class tool exists for an action, use the tool
directly.

## Tool Schemas
Complete JSON Schema for available tools:
- {tool_name}
```json
{json_schema}
```
- ...
\end{Verbatim}

\begin{Verbatim}[frame=single,rulecolor=\color{ysdarkblue},fillcolor=\color{ysshallowblue},label=\textbf{OpenClaw System Prompt -- Behavior Rules},fontsize=\small]
## Tool Call Style
Default: do not narrate routine, low-risk tool calls (just
call the tool).
Narrate only when it helps: multi-step work, complex tasks,
or sensitive actions. Keep narration brief and value-dense.
Tool-call protocol is strict: use native API tool/function
calls only. Never emit tool calls as plain text markup
(e.g. <tool_call>, <function=...>, <parameter=...>). If a
tool is needed, issue a real tool call instead of describing
or simulating it in text.

## Safety
- Safety: {safety_rule}
- Tool Call Style: {tool_call_style_rule}
- Reply Tags: {reply_tags_rule}
- Silent Reply: {silent_reply_rule}
- Heartbeat: {heartbeat_rule}
\end{Verbatim}

\begin{Verbatim}[frame=single,rulecolor=\color{ysdarkblue},fillcolor=\color{ysshallowblue},label=\textbf{OpenClaw System Prompt -- Skills Block (template)},fontsize=\small]
## Skills (mandatory)
Before replying: scan <available_skills> entries.
- If exactly one skill clearly applies: read its SKILL.md
  using the configured read tool, then follow it.
- If multiple skills could apply: choose the most specific
  one, then read and follow it.
- If none clearly apply: do not read any SKILL.md.
Constraints: never read more than one skill up front; only
read after selecting.

<available_skills>
  <skill>
    <name>{skill_name}</name>
    <description>{skill_description}</description>
    <location>{skill_path}</location>
  </skill>
  ...
</available_skills>
\end{Verbatim}

The four workspace files (\texttt{AGENTS.md}, \texttt{SOUL.md}, \texttt{USER.md}, \texttt{TOOLS.md}) are injected verbatim under a \texttt{\#\# Workspace Files} header. Their contents are task-specific and define the agent's persona, the simulated user's profile, and any additional tooling the task is meant to follow; we omit them for brevity.

\subsection{CLAW-Eval: simulated user agent (multi-turn split)}
\label{app:agent_claw_user}

The \textit{multi-turn} split of CLAW-Eval evaluates dialogue between the agent and a human user. Rather than scripting fixed user turns, OpenClaw runs a second LLM as a \textit{user agent} that produces user replies on the fly, conditioned on a per-task persona. We use Claude Opus~4.6 in this role. The system prompt below is an English rendering of the upstream prompt (the original is in Chinese in the OpenClaw codebase); the verbatim Chinese text is included with the released code.

\begin{Verbatim}[frame=single,rulecolor=\color{ysdarkblue},fillcolor=\color{ysshallowblue},label=\textbf{Simulated User Agent -- System Prompt (translated)},fontsize=\small]
You are a simulated user. Hold a conversation with an AI
assistant according to the persona below.

## Your persona
{persona}

## Rules
1. Stay in character. Reply in natural, spoken-style language;
   do not reveal that you are an AI.
2. Answer the assistant's questions truthfully based on your
   persona.
3. If the assistant asks for information not in your persona,
   say something like "I don't really know the exact number"
   or a similar natural reply.
4. If the assistant has already given a complete answer or
   recommendation and you have no further questions, output
   [DONE].
5. If you are satisfied with the assistant's reply, or it has
   fully addressed your question, output [DONE].
6. Keep replies short and natural, like a real user (1-3
   sentences).
\end{Verbatim}

At each turn, the user-agent LLM is given (i) the rendered persona, (ii) the running transcript of agent$\leftrightarrow$user messages, and (iii) an instruction to reply or to emit \texttt{[DONE]} if satisfied. A \texttt{[DONE]} reply ends the dialogue and triggers grading.

\section{Baseline Configurations}
\label{app:baselines}

We give the full configuration of the four baselines introduced in Section~\ref{sec:experiments}.

\subsection{\textsc{Reminder}}
\label{app:reminder_prompt}

At every agent turn, we append the following text to the user message. The wording is fixed across all turns and tasks.

\begin{Verbatim}[frame=single,rulecolor=\color{ysdarkblue},fillcolor=\color{ysshallowblue},label=\textbf{\textsc{Reminder} -- Per-Turn Suffix},fontsize=\small]
[Reminder] Don't overthink: if a single tool call would
settle a question, call the tool instead of deliberating.
Don't overact: if you just received an observation, reason
about it before issuing the next command.
\end{Verbatim}

\subsection{\textsc{AgentReflect}}
\label{app:agentreflect_prompt}

Every $K{=}10$ steps we pause the agent and invoke a reflector LLM (same backbone as the agent) on the last 10 steps plus the running context. The reflector emits a fixed three-line template, prepended as a system message to the next turn. We set $K{=}10$ to match the judge's chunk size (Section~\ref{sec:judge}).

\begin{Verbatim}[frame=single,rulecolor=\color{ysdarkblue},fillcolor=\color{ysshallowblue},label=\textbf{\textsc{AgentReflect} -- Reflector System Prompt},fontsize=\small]
You are a reviewer for a coding agent's recent steps.
After reading the last 10 steps and the running task
context, output exactly three lines:

1. Progress: one sentence on the most useful new
   information the agent obtained in this chunk.
2. Stuck signals: one sentence flagging any sign of
   being stuck (e.g., re-reading the same file,
   repeating a command, circling between the same
   two hypotheses). Write "none" if nothing stands out.
3. Next direction: one sentence recommending the single
   most useful next action.

Keep each line to one sentence. Do not include any other
text.
\end{Verbatim}

\subsection{\textsc{ReCAP} and \textsc{AgentDebug}}
\label{app:external_baselines}

\textsc{ReCAP}~\citep{recap2025}: we follow the upstream protocol, re-injecting the parent plan into the active prompt at each recursive return.
\textsc{AgentDebug}~\citep{zhang2025agentdebug}: on failed rollouts, the upstream root-cause decomposer locates the first faulty step and emits verbal feedback; we prepend it to the prompt for a retry, and report that retry's resolve rate.

\section{LLM-as-Judge Prompt Templates}
\label{app:prompt}

Labeling each step requires knowing what the agent has already learned, so the judge must track accumulated context across the full trajectory.
We provide the full system prompts for the three-stage pipeline described in Section~\ref{sec:judge}.

The pipeline works as follows.
(1) The \textbf{State Maintainer} reads the raw steps and updates a structured rolling state with two components: KnownFacts (observable facts) and Progress (progress assessment).
(2) The \textbf{State Validator} audits the KnownFacts component against the raw steps, flagging hallucinations, omissions, and contradictions; if errors are found, the maintainer revises and the validator re-audits in a reflection loop (up to 3 iterations).
(3) The \textbf{Annotator} labels each step relative to the validated rolling state, assigning one of three labels along with a confidence level and a one-sentence justification.

\subsection{Stage 1: State Maintainer}
\label{app:state_maintainer}

\begin{Verbatim}[frame=single,rulecolor=\color{ysdarkblue},fillcolor=\color{ysshallowblue},label=\textbf{State Maintainer -- System Prompt},fontsize=\small]
You are a trajectory state tracker. Your job is to read a
chunk of coding agent steps and update the rolling state to
accurately reflect what the agent has learned and done.

You are NOT judging the agent's decisions. You are tracking
two things so a downstream annotator can decide whether each
step grows what the agent knows:

1. **KnownFacts** -- observable facts only:
   - What files were read and what was learned from them
   - What hypothesis the agent is currently working with
   - What edits were made
   - What test results were observed
   - Critical discoveries

2. **Progress** -- whether KnownFacts is growing:
   - What phase the agent is in
   - steps_since_progress: number of steps since
     KnownFacts last grew
   - repetition_tracker: signs of effort without growth
     (files re-read, repeated action patterns, recurring
     hypothesis cycles)

Be precise and factual. Do not speculate beyond what the
steps show.
\end{Verbatim}

The user message includes four sections: (1) Agent's Task Setup (the GitHub issue description), (2) Current Rolling State as formatted JSON through the previous chunk, (3) New Steps for the current chunk with per-step thinking, action, and observation fields, and (4) a task instruction requesting only valid JSON output matching the rolling state schema.

\subsection{Stage 2: State Validator}
\label{app:state_validator}

\begin{Verbatim}[frame=single,rulecolor=\color{ysdarkblue},fillcolor=\color{ysshallowblue},label=\textbf{State Validator -- System Prompt},fontsize=\small]
You are a FACTUAL auditor for a coding-agent trajectory's
KnownFacts state. You receive a proposed KnownFacts object
and the raw trajectory steps it was derived from. Your ONLY
job is to detect factual errors in the KnownFacts that
contradict the raw steps.

You are NOT auditing phase, progress, or repetition -- only
the KnownFacts fields (files_read, hypotheses, edits_made,
test_results, key_findings).

Report ONLY these three error types:

1. HALLUCINATION -- KnownFacts claims something that does NOT
   appear in the raw steps. Examples:
     - claims a file was read, but no step reads that file
     - claims a test passed, but the observation shows failure
     - claims an edit was made, but no edit action occurred

2. OMISSION -- a concrete, observable fact IS in the raw steps
   but is missing from KnownFacts. Examples:
     - an edit was committed but edits_made is empty
     - a test was run with a clear pass/fail result but
       test_results does not mention it

3. CONTRADICTION -- KnownFacts directly contradicts text that
   is verbatim visible in a step's observation.

DO NOT report any of the following as corrections:
  - Paraphrasing, rewording, or stylistic differences
  - Disagreement over subjective framing (hypothesis wording,
    whether a finding is "key")
  - Summaries that are less detailed than you would write
    -- shorter is fine
  - Anything that requires guessing what the agent "meant"

If you find none of (1)(2)(3), you MUST return
"corrections": [] and copy the proposed KnownFacts into
validated_known_facts unchanged.
\end{Verbatim}

The validator responds with structured JSON containing a list of corrections (each prefixed with the error type) and a \texttt{validated\_known\_facts} object.
When corrections are found, the State Maintainer receives the error list and revises the rolling state; the validator re-audits the revision in a reflection loop (up to 3 iterations).

\subsection{Stage 3: Annotator}
\label{app:annotator}

\begin{Verbatim}[frame=single,rulecolor=\color{ysdarkblue},fillcolor=\color{ysshallowblue},label=\textbf{Annotator -- System Prompt},fontsize=\small]
You are an expert judge evaluating the epistemic efficiency
of a coding agent. Each step is one of: OVERTHINKING (OT),
OVERACTING (OA), CALIBRATED (CAL).

A step s_t = (reasoning r_t, action a_t, observation o_t) is
judged by what it contributes to the agent's known facts
BEFORE that step (KnownFacts below).

## Definitions

- **CALIBRATED** -- the step expands KnownFacts. Either r_t
  derives a new hypothesis or finding from prior facts, or
  a_t / o_t adds a file, command result, test outcome, or
  edit not yet in KnownFacts.

- **OVERTHINKING** -- r_t re-derives a conclusion already in
  KnownFacts, cycles between known hypotheses, restates
  known information, or deliberates about something a single
  tool call would settle. Internal tokens are spent, but
  KnownFacts does not grow.

- **OVERACTING** -- a_t returns an observation already in
  KnownFacts (e.g. re-reads a file already in files_read,
  re-runs a test already in test_results without an
  intervening edit, repeats a prior command), or fires
  without reasoning over information just received. A tool
  call is spent, but KnownFacts does not grow.

## Rules

- Judge each step against KnownFacts as it stood BEFORE
  that step.
- The rolling state is a compressed REFERENCE for what the
  agent knew; if it and the raw steps disagree, trust the
  raw steps.
- If a step contains more than one failure mode (e.g.,
  redundant reasoning followed by a redundant tool call),
  assign the label corresponding to the dominant failure.
  The dominance order is OT > OA > CAL.
- When uncertain, mark confidence as "low".
\end{Verbatim}

The annotator outputs structured JSON with one entry per step, each containing a label (\texttt{OVERTHINKING}, \texttt{OVERACTING}, or \texttt{CALIBRATED}), a one-sentence reasoning, and a confidence level (\texttt{high}, \texttt{medium}, or \texttt{low}). The OT $>$ OA $>$ CAL priority is motivated by the temporal structure of a step: reasoning occurs before action, so once $r_t$ is classified as overthinking, the step is already non-progressive with respect to expanding $\mathcal{K}_t$, independent of the subsequent action $a_t$.

\subsection{Rolling State Schema}
\label{app:rolling_state}

The rolling state is the central data structure that carries context across chunks.
It has two components, each serving a distinct role in the pipeline.

The \texttt{known\_facts} component stores observable facts extracted from the trajectory.
\texttt{files\_read} maps each filename to a one-sentence summary of what the agent learned from it.
\texttt{hypotheses} records the agent's current working theory about the bug.
\texttt{edits\_made} and \texttt{test\_results} track what the agent changed and what feedback it received.
\texttt{key\_findings} captures critical discoveries that shape subsequent decisions.
The State Validator (Section~\ref{app:state_validator}) audits only this component, checking each field against the raw steps for hallucinations, omissions, and contradictions.

The \texttt{progress} component tracks the agent's behavioral trajectory.
\texttt{phase} classifies the agent's current activity into one of six stages: \texttt{understanding\_issue}, \texttt{localizing\_code}, \texttt{editing}, \texttt{testing}, \texttt{verifying}, or \texttt{stuck\_looping}.
\texttt{progress\_trend} maintains a per-chunk log of whether the agent advanced toward a solution.
\texttt{steps\_since\_progress} counts consecutive steps without meaningful progress.
\texttt{repetition\_tracker} records three types of redundancy: files read more than once, repeated action patterns (e.g., running the same grep command), and circular reasoning cycles where the agent revisits hypotheses it already explored.
The Annotator uses this component to distinguish calibrated behavior from overthinking (circular reasoning, stalled progress) and overacting (redundant file reads, repeated commands).

\begin{Verbatim}[frame=single,rulecolor=\color{ysdarkblue},fillcolor=\color{ysshallowblue},label=\textbf{Rolling State JSON Schema},fontsize=\small]
{
  "known_facts": {
    "files_read": {
      "<filename>": "<what was learned -- one sentence>"
    },
    "hypotheses": "<current working hypothesis>",
    "edits_made": "<what was changed and where>",
    "test_results": "<which tests ran and their outcome>",
    "key_findings": "<critical discoveries>"
  },
  "progress": {
    "phase": "<one of: understanding_issue,
      localizing_code, editing, testing,
      verifying, stuck_looping>",
    "progress_trend": [
      "<chunk-level progress assessment>"
    ],
    "steps_since_progress": <int>,
    "repetition_tracker": {
      "files_read_multiple": ["<filenames>"],
      "repeated_patterns": ["<action patterns>"],
      "circular_reasoning": ["<hypothesis cycles>"]
    }
  }
}
\end{Verbatim}

We show a concrete example below.
The state comes from the first chunk (steps 0--9) of a trajectory on \texttt{django-11163}, where the agent must fix \texttt{model\_to\_dict()} treating \texttt{fields=[]} as falsy.
By step 9, the agent has read the source, identified the root cause, applied a one-line fix, and verified the output.
The \texttt{known\_facts} component captures these facts; the \texttt{progress} component records that the agent is in the verifying phase, has re-read \texttt{models.py} multiple times, and has circled back to the same explanation of the bug.

\begin{Verbatim}[frame=single,rulecolor=\color{ysdarkblue},fillcolor=\color{ysshallowblue},label=\textbf{Rolling State Example (django-11163, after steps 0--9)},fontsize=\small]
{
  "known_facts": {
    "files_read": {
      "django/forms/models.py": "model_to_dict() used
        `if fields and f.name not in fields:` while
        construct_instance() already used
        `if fields is not None and ...`.",
      "tests/runtests.py": "Django's test settings are
        configured dynamically rather than via a single
        tests.settings module."
    },
    "hypotheses": "The bug is caused by model_to_dict()
      treating an empty list for `fields` as falsy, so
      changing the condition to `fields is not None`
      should fix it.",
    "edits_made": "Modified model_to_dict() by changing
      `if fields and f.name not in fields:` to
      `if fields is not None and f.name not in fields:`.",
    "test_results": "Reproduction script confirmed
      model_to_dict(instance, fields=[]) returned all
      fields before the fix and {} after the fix.",
    "key_findings": "The same `fields is not None` pattern
      already existed in construct_instance(); `exclude`
      checks did not require the same change."
  },
  "progress": {
    "phase": "verifying",
    "progress_trend": [
      "Localized the bug in django/forms/models.py.",
      "Reproduced the issue, applied the fix, and
       verified the expected outputs."
    ],
    "steps_since_progress": 0,
    "repetition_tracker": {
      "files_read_multiple": ["django/forms/models.py"],
      "repeated_patterns": [
        "Tried multiple environment setups for the
         reproduction script after import errors."
      ],
      "circular_reasoning": [
        "Repeated the same explanation that falsy
         fields=[] bypasses the filter."
      ]
    }
  }
}
\end{Verbatim}


\end{document}